%% file: 0_cvpr_main.tex
\documentclass[final]{cvpr}

\usepackage{times}
\usepackage{epsfig}
\usepackage{graphicx}
\usepackage{subcaption}

\usepackage{abstract}

\usepackage{amsmath}
\usepackage{amssymb}
\usepackage[dvipsnames]{xcolor}
\usepackage{etoolbox}
\usepackage{capt-of,etoolbox}
\usepackage{enumitem}
\usepackage{multirow}
\setlist{nosep}
\usepackage[ruled,vlined,linesnumbered]{algorithm2e}
\SetInd{4pt}{4pt}
\DontPrintSemicolon
\SetKwProg{Def}{def}{:}{}

\usepackage[pagebackref=true,breaklinks=true,colorlinks,bookmarks=false]{hyperref}

\def\cvprPaperID{1924} % *** Enter the CVPR Paper ID here
\def\confYear{CVPR 2021}

\newcommand{\ScoreSum}{Q}

\newcommand{\calI}{\mathcal{I}}
\newcommand{\calC}{\mathcal{C}}

\newcommand{\calN}{\mathcal{N}}
\newcommand{\calO}{\mathcal{O}}
\newcommand{\calP}{\mathcal{P}}

\newcommand{\curr}{\text{curr}}

\newcommand{\pool}{\text{pool}}

\newcommand*\samethanks[1][\value{footnote}]{\footnotemark[#1]}

\newcommand{\centercell}[2]{\multirow{2}{*}{\parbox{#1}{\centering #2}}}

\begin{document}

\title{Monte Carlo Scene Search for 3D Scene Understanding}

\author{\thanks{The first two authors contributed equally.}~ Shreyas Hampali\textsuperscript{(1)}, \samethanks~ Sinisa Stekovic\textsuperscript{(1)},
Sayan Deb Sarkar\textsuperscript{(1)}, Chetan S. Kumar\textsuperscript{(1)}, \\ Friedrich Fraundorfer\textsuperscript{(1)}, 
Vincent Lepetit\textsuperscript{(2,1)} \and
\textsuperscript{(1)}{\normalsize Institute for Computer Graphics and Vision, Graz University of Technology, Graz, Austria }\\ 
\textsuperscript{(2)}{\normalsize Universit\'e Paris-Est, \'Ecole des Ponts ParisTech, Paris, France} \\
{\tt\small \{<firstname>.<lastname>\}@icg.tugraz.at, fraundorfer@icg.tugraz.at, vincent.lepetit@enpc.fr} \\
{\small Project page: \href{https://www.tugraz.at/index.php?id=50484}{ \color{blue} https://www.tugraz.at/index.php?id=50484}}
}

\twocolumn[{%
\renewcommand\twocolumn[1][]{#1}%
\maketitle

\begin{center}
%% \centering
\includegraphics[trim=0 45 0 20,clip,width=.83\linewidth]{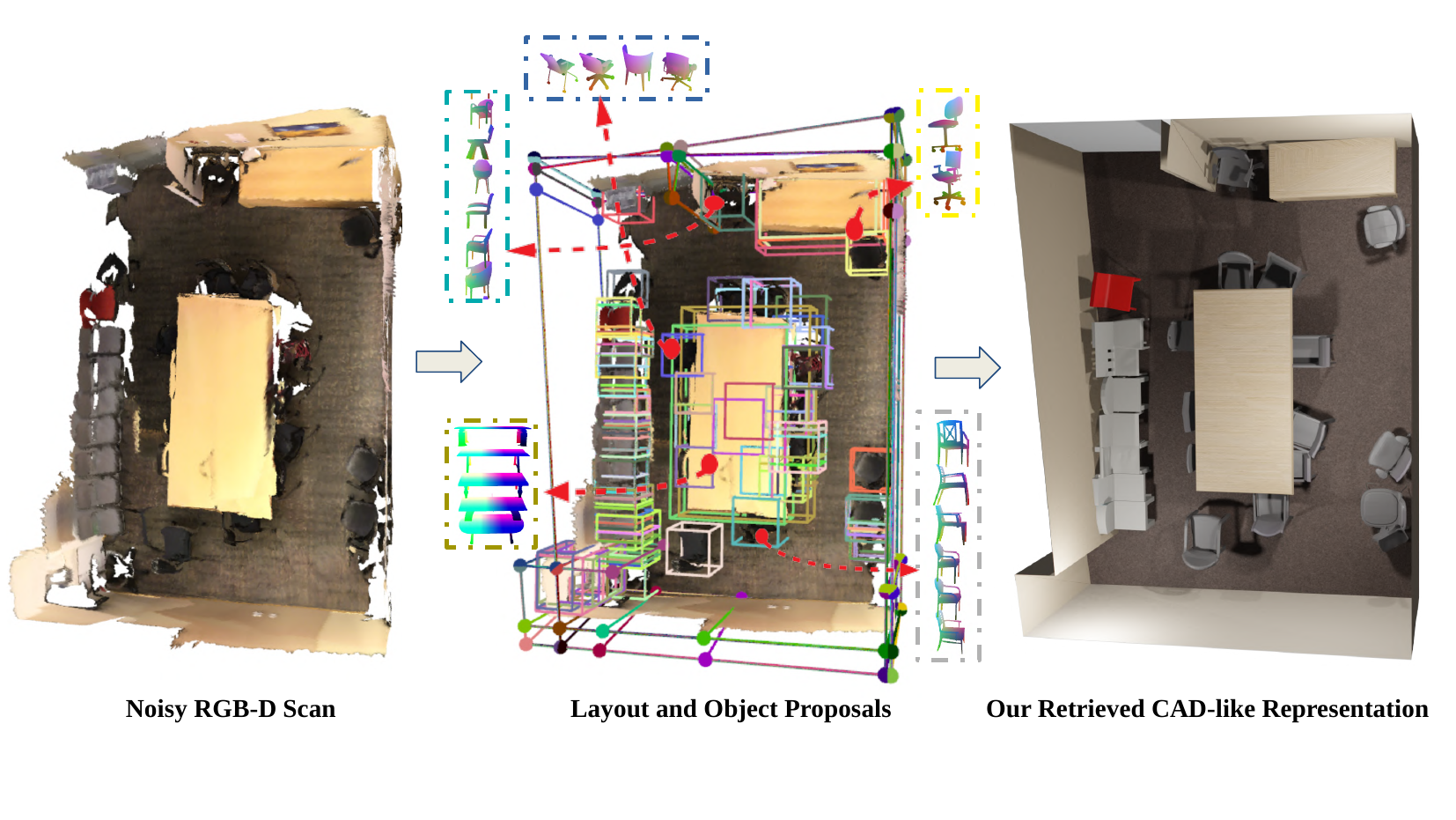}
\vspace{-0.3cm} 
\captionof{figure}{In this paper, we advocate for the use of Monte Carlo Tree Search~(MCTS) for 3D scene understanding problems. Given a noisy 3D point cloud recovered from an RGB-D sequence, our approach recovers accurate 3D models and poses for the objects, walls, and floor with minimal training data, even in challenging conditions.  We first generate proposals for the layout components and the objects, and rely on Monte Carlo Tree Search~(MCTS) adapted to the problem to identify the proposals that best explain the RGB-D sequence.  We retrieve correctly the arrangement of chairs on the left-hand side of the scene despite them being close to each other and the thin wall on the top. Our adapted MCTS algorithm has few hyperparameters and can be applied to wide variety of scenes with minimal tuning effort.
For visualization purposes only, we texture the objects and the layout using the colors of the 3D points close-by.} \label{fig:teaser}
\end{center}
}]

%%%%%%%%% ABSTRACT
\begin{abstract}
We explore how a general AI algorithm can be used for 3D scene understanding to reduce the need for training data. More exactly, we propose a modification of the Monte Carlo Tree Search~(MCTS) algorithm to retrieve objects and room layouts from noisy RGB-D scans.  While MCTS was developed as a game-playing algorithm, we show it can also be used for complex perception problems.  Our adapted MCTS algorithm has few easy-to-tune hyperparameters and can optimise general losses. We use it to optimise the posterior probability of objects and room layout hypotheses given the RGB-D data.  This results in an analysis-by-synthesis approach that explores the solution space by rendering the current solution and comparing it to the RGB-D observations. To perform this exploration even more efficiently, we propose simple changes to the standard MCTS' tree construction and exploration policy.  We demonstrate our approach on the ScanNet dataset.  Our method often retrieves configurations that are better than some manual annotations, especially on layouts.
\end{abstract}

% , indicating that it could be used to automatically generate 3D annotations to train Deep Learning approaches for faster inference

%%%%%%%%% BODY TEXT

\saythanks

\input{Intro}

\input{RelatedWork}

\input{MCTS_Overview}

\input{Method}

\input{Evaluations}

% \input{AblationStudy}
% \section{Acknowledgments}
 \noindent\textbf{Acknowledgments.} This work  was supported by the Christian Doppler Laboratory for Semantic 3D Computer Vision, funded in part by Qualcomm Inc.

% \newpage
{\small
\bibliographystyle{ieee_fullname}
\bibliography{egbib}
}

\end{document}

\title{Monte Carlo Scene Search for 3D Scene Understanding\\ Supplementary Material}

\author{\thanks{The first two authors contributed equally.}~ Shreyas Hampali\textsuperscript{(1)}, \samethanks~ Sinisa Stekovic\textsuperscript{(1)},
Sayan Deb Sarkar\textsuperscript{(1)}, Chetan S. Kumar\textsuperscript{(1)}, \\ Friedrich Fraundorfer\textsuperscript{(1)}, 
Vincent Lepetit\textsuperscript{(2,1)} \and
\textsuperscript{(1)}Institute for Computer Graphics and Vision, Graz University of Technology, Graz, Austria \\ 
\textsuperscript{(2)}Universit\'e Paris-Est, \'Ecole des Ponts ParisTech, Paris, France \\
}

\maketitle

\input{Supplementary}

% \newpage~\newpage
{\small
\bibliographystyle{ieee_fullname}
\bibliography{egbib}
}

%% file: Intro.tex
\section{Introduction}

3D scene understanding is a fundamental  problem in Computer Vision~\cite{Roberts65,Yakimovsky73}. In the case of indoor scenes, one usually aims at recognizing the objects and their properties such as their 3D pose and geometry~\cite{Avetisyan_2019_CVPR,Avetisyan2019EndtoEndCM,Grabner20183DPE}, or the room layouts~\cite{zhang2015large,liu2017raster,Zou20193d,Structured3D,liu2018floornet, Pintore2020AtlantaNetIT, Sun2019horizonnet, Zou18, Zou20193d, zeng2020joint, zhang2020geolayout}, or both~\cite{avetisyan2020scenecad,huang2018cooperative,Nie_2020_CVPR,shao2014imagining,Tulsiani17,Zhang}.  With the development of deep learning approaches, the field has made a remarkable progress. Unfortunately, all recent methods are trained in a supervised way on 3D annotated data. Such a supervised approach has several drawbacks: 3D manual annotations are particularly cumbersome to create and creating realistic virtual 3D scenes also has a high cost~\cite{Roberts20}. Moreover, supervised methods also tend to generalize poorly to other datasets. Even more importantly, they can only be as good as the training 3D annotations, and mistakes in manual annotations are actually common in existing datasets, as we will show. If one wants to go further and consider more scenes without creating real or synthetic training datasets, it seems important to be able to develop methods that do not rely too much on 3D scenes for training.

Over the history of 3D scene understanding, many non-supervised approaches have already been proposed, including recently to leverage deep learning object detection methods. They typically combine generative models and the optimization of their parameters. Generative methods for 3D scene understanding indeed often involve optimization problems with high complexity, and many optimization tools have thus been investigated, including Markov Random Fields~(MRFs) and Conditional Random Fields~(CRFs)~\cite{Koppula11,Wang15,mura2016}, Markov Chains Monte Carlo~(MCMCs)~\cite{Chen2019HolisticSU, huang2018holistic, choi, Zhao}, tree search~\cite{Lee2010EstimatingSL}, or hill climbing~\cite{ZouGLH19, izadinia2017im2cad}. However, there does not seem to be a clear method of choice:  MRFs and CRFs impose strong constraints on the objective function; MCMCs depend on many hyperparameters that are difficult to tune and can result in slow convergence; hill climbing can easily get stuck in a local optimum. The tree search method used by \cite{Lee2010EstimatingSL} uses a fixed width search tree that can miss good solutions.

% In the absence of supervision for guidance, this game can become a very complex task in an open world: The search space has a size of $2^N$, where $N$ has an order of magnitude of 100 to 1000, for scenes such as the one depicted in Fig.~\ref{fig:teaser}. 

In this paper, we advocate for the use of Monte Carlo Tree Search~(MCTS)~\cite{Coulom06,browne2012survey}, which is a general discrete AI algorithm for learning to play games~\cite{AlphaZero}, for optimization in 3D scene understanding problems. We propose to see perception as a (single-player) game, where the goal is to identify the right 3D elements that explain the scene.    In such cases where the search problem can be organized into a tree structure which is too large for exhaustive evaluation, MCTS becomes a very attractive option. It also depends on very few easy-to-tune hyperparameters. Moreover, it can be interrupted at any time to return the best solution found so far, which can be useful for robotics applications. A parallel implementation is also possible for high efficiency~\cite{Chaslot08b}. In short, MCTS is a powerful optimization algorithm, but to the best of our knowledge, it has never been applied to 3D perception problems.

% It is, for example, a key component in AlphaGo and AlphaZero~\cite{AlphaZero}, an algorithm that achieves super-human performance for several two-player games, such as Go and chess.

To apply MCTS to 3D scene understanding, as shown in Fig.~\ref{fig:teaser}, we generate proposals for possible objects and layout components using the point cloud generated from the RGB-D sequence, as previous works do from a single RGB-D frame~\cite{Lee2010EstimatingSL,ZouGLH19}.  MCTS can be used to optimize general loss functions, which do not even have to be differentiable.  This allows us to rely on a loss function based on an analysis-by-synthesis (or ``render-and-compare'') approach to select the proposals that correspond best to the observations. Our loss function compares (non-realistic) renderings of a set of proposals to the input images and can incorporate constraints between the proposals. This turns MCTS into an analysis-by-synthesis method that explores possible sets of proposals for the observations, possibly back-tracking to better solutions when an exploration does not appear promising.

We adapted the original MCTS algorithm to the 3D scene understanding problem to guide it towards the correct solution faster, and call the resulting method ``MCSS'', for \textit{Monte Carlo Scene Search}.  First, it is possible to structure the search tree so that it does not contain any impossible solutions, for example, solutions with intersecting proposals.  We also enforce the exploration of proposals which are close spatially to proposals in the same path to the root node.  Second, we introduce a score based on how the proposal improves the solution locally to increase the efficiency of search.% balance exploration and exploitation in the MCTS algorithm.

In practice, we first run MCSS only on the layout proposals to recover the layout. We then run MCSS on the object proposals using the recovered layout. The recovery of the objects thus exploits constraints from the layout, which we found useful as shown in our experiments. In principle, it is possible to run a single MCSS on both the object and layout component proposals, but constraints from the objects did not appear useful to constrain the recovery of the layout for the scenes in ScanNet, which we use to evaluate our approach. We therefore used this two-step approach for simplicity. It is, however, possible that more complex scenes would benefit from a single MCSS running on all the proposals.

Running our method takes a few minutes per scene. This is the same order of magnitude as the time required to acquire an RGB-D sequence covering the scene, but definitively slower than supervised methods. However, our direction could lead to a solution that automatically generates annotations, which could be used to train supervised methods for fast inference. We show in the experiments that our method already retrieves annotations that are sometimes more accurate than existing manual annotations, and that it can be applied to new data without tuning any parameters.  Beyond that,  MCTS is a very general algorithm, and the approach we propose could be transposed to other perception problems and even lead to an integrated architecture between perception and control, as MCTS has also already been applied to robot motion planning control~\cite{labbe20}.

%% file: RelatedWork.tex
\section{Related Work}
3D scene understanding is an extremely vast topic of the computer vision literature.  We focus here on indoor layout and object recovery, as we demonstrate our approach on this specific problem.

% The scenes we reconstruct consist of layout components and objects and thus we discuss the recent work in layout estimation, object retrieval and methods which consider both problems jointly. 

% \vincentrmk{Could you precise more the type of input? single image,
%   single RGB-D, point cloud?}
%  \shreyasrmk{Fixed for 2.2 and 2.3. Sinisa can you look at 2.1?}

\subsection{Layout Estimation}

The goal of layout estimation is to recover the walls, floor(s), and ceiling(s) of a room or several rooms. This can be very challenging as layout components are often partially or completely occluded by furniture.  Hence, many methods resort to some type of prior or supervised learning.  The cuboid assumption constraints the room layout to be a box~\cite{Schwing12, Hedau09, lee2017roomnet}.  The Manhattan assumption relaxes somewhat this prior, and enforces the components to be orthogonal or parallel.  Many methods working from panoramic images~\cite{Sun2019horizonnet, Zou18, Zou20193d} and point clouds~\cite{ikehata2015structured, murali2017indoor, sanchez2012planar} rely on such priors.  
Methods which utilize supervised learning~\cite{zhang2015large,liu2017raster,Zou20193d,Structured3D,liu2018floornet, Pintore2020AtlantaNetIT, Sun2019horizonnet, Zou18, Zou20193d, zeng2020joint, zhang2020geolayout} depend on large-scale datasets, the creation of which is a challenge on its own.
When performing layout estimation from point clouds as input data~\cite{sanchez2012planar, cabral2014piecewise, ikehata2015structured, murali2017indoor, mura2016}, one has to deal with incomplete and noisy scans as can be found in the ScanNet dataset~\cite{dai2017scannet}. Like previous work~\cite{murali2017indoor,stekovic2020general}, we  first hypothesize layout component proposals, but relying on MCTS for optimization lets us deal with a large number of proposals and be robust to noise and missing data, without special constraints like the Manhattan assumption.

\subsection{3D Object Detection and Model Retrieval}
Relevant to our work are techniques to detect objects in the input data and to predict their 3D pose and the 3D model. If 3D data is available, as in our case, this is usually done by first predicting 3D bounding boxes from RGB-D~\cite{Lin13,Song2014SlidingSF,Song7780463} or point cloud data~\cite{qi2019deep,hou2019sis, Qi8578200, Qi2020ImVoteNetB3, Song7780463} as input. One popular way to retrieve the geometry of objects from indoor point clouds is to predict an embedding and retrieve a CAD model from a database~\cite{Avetisyan_2019_CVPR,Avetisyan2019EndtoEndCM,dahnert2019embedding, Grabner20183DPE, Kuo2020Mask2CAD3S}.

However, while 3D object category detection and pose estimation from images is difficult due to large variations in appearance, it is also challenging with RGB-D scans due to incomplete depth data. 
Moreover, in cluttered scenarios, it is still difficult to get all the objects correctly~\cite{Kulkarni20193DRelNetJO}. To be robust, our approach generates many 3D bounding box proposals and multiple 
possible CAD models for each bounding box. We then rely on MCTS to obtain the optimal combination of CAD models which fits the scene.

\subsection{Complete scene reconstruction}
Methods for complete scene reconstruction consider both layout and objects. Previous methods fall into two main categories, generative and discriminative methods. 

Generative methods often rely on an analysis-by-synthesis approach. A recent example for this is \cite{izadinia2017im2cad} in which the room layout (under cuboid assumption) and alignment of the objects are optimized using a hill-climbing method. 
Some methods rely on a parse graph as a prior on the underlying structure of the scene~\cite{Chen2019HolisticSU, huang2018holistic, choi, Zhao}, and rely on a stochastic Markov Chain Monte Carlo~(MCMC) method to find the optimal structure of the parse graph and the component parameters. Such a prior can be very useful to retrieve the correct configuration, unfortunately MCMCs can be difficult to tune so that they work well on all scenes with the same parameters.

Like us, other works deal with an unstructured list of proposals~\cite{Lee2010EstimatingSL,ZouGLH19}, and search for an optimal set which minimizes a fitting cost defined on the RGB-D data. Finding the optimal configuration of components constitutes a subset selection problem. In~\cite{ZouGLH19},  due to its complexity, it is solved using a greedy hill-climbing search algorithm. In~\cite{Lee2010EstimatingSL}, it is solved using beam search on the generated hypothesis tree with a fixed width for efficiency, which can miss good solutions in complex cases. 
Our approach is similar to~\cite{Lee2010EstimatingSL, ZouGLH19} as we also first generate proposals and aim at selecting the correct ones, but for the exploration of the search tree, we propose to utilize a variant of Monte Carlo Tree Search, which is known to work well even for very large trees thanks to a guided sampling of the tree.

Discriminative methods can exploit large training datasets to learn to classify scene components from input data such as RGB and RGB-D images~\cite{avetisyan2020scenecad,huang2018cooperative,Nie_2020_CVPR,Tulsiani17,Zhang}. By introducing clever Deep Learning architectures applied to point clouds or voxel-based representations, these methods can achieve very good results. However, supervised methods have practical drawbacks: They are limited by the accuracy of the annotations on which they are trained, and high-quality 3D annotations are difficult to create in practice; generalizing to new data outside the dataset is also challenging. In the experiments, we show that without any manually annotated data, our method can retrieve accurate 3D scene configurations on both ScanNet and our own captures even for cluttered scenes, and with the same hyperparameters.

%% file: MCTS_Overview.tex
\section{Overview of MCTS}

For the sake of completeness, we provide here a brief overview of MCTS.  An in-depth survey can be found in~\cite{browne2012survey}.  MCTS solves problems of high complexity that can be formalized as tree search  by sampling paths throughout the tree and evaluating their scores.  Starting from a tree only containing  the root node, this tree is gradually expanded in the most promising directions.  To identify the most promising solutions (\ie paths from the root node to a leaf node), a score for each created node is evaluated through ``simulations'' of complete games.  A traversal starting from a node can choose to continue with an already visited node with a high score~(exploitation) or to try a new node~(exploration).  MCTS performs a large number of tree traversals, each starting from the root node following four consecutive phases we describe below.  The pseudo-code for single-player non-random MCTS, which corresponds to our problem, is given in the supplementary material.

%% Algorithm~\ref{alg:mcts}.

%% \input{MCTS_Algo}

\noindent {\bf $\textsc{Select}$. } This step selects the next node of the tree to traverse among the children of the current node $\calN_\curr$. (case 1) If one or several children have not been visited yet, one of them is selected randomly and MCTS moves to the $\textsc{Expand}$ step. (case 2) If all the children have been visited at least once, the next node is selected based on some criterion. The most popular criterion to balance exploitation and exploration is the Upper Confidence Bound~(UCB)~\cite{Auer02}:
  \begin{align}
  \arg\displaystyle\max_{\calN \in \calC(\calN_\curr)} \lambda_1 \frac{\ScoreSum(\calN)}{n(\calN)} +
  \lambda_2 \cdot \sqrt{\frac{\log n(\calN_{curr})}{n(\calN)}} \> ,
  \label{eq:UCB}
\end{align}
  where $\calC(\calN_\curr)$ is the set of children nodes for the current node, $Q(\calN)$ is a sum of scores obtained through simulations, and $n(\calN)$ is the number of times $\calN$ is traversed during the search.  The selected node is assigned to $\calN_\curr$, before iterating the $\textsc{Select}$ step.  Note that in single-player games, the maximum score is sometimes used in place of the average for the first term, as there is less uncertainty.  We tried both options and they perform similarly in our case.
 
%% \vincent{Note that two-player games usually consider the average value rather than the maximum, which is more adapted to one-player games as in our case. This is because we do not need to account for the uncertainty of a the opponent's moves in single-player games, which makes the game score more reliable.}  \vincentrmk{are the two last sentence correct?}\sinisarmk{We should remove this paragraph. You can see in the equation $\ref{eq:mcts_criterion}$ we also use average now.}

\noindent {\bf $\textsc{Expand}$. } In case 1, this step expands the tree by adding the randomly selected node to the tree.

\noindent {\bf $\textsc{Simulate}$.} After the $\textsc{Expand}$ step, many ``simulations'' of the game are run  to assign the new node $\calN$ a score, stored in $Q(\calN)$.  Each simulation follows a randomly-chosen path from the new node until the end of the game. The score can be for example the highest score obtained by a simulation at the end of the game.

\noindent {\bf $\textsc{Update}$.} After the $\textsc{Simulate}$ step, the score is also added to the $Q$ values of the ancestors of $\calN$.
The next MCTS iteration will then traverse the tree from the root node using the updated scores.

After a chosen number of iterations, in the case of non-random single-player games, the solution returned by the algorithm is the simulation that obtained the best score for the game.

%% ~\cite{bjornsson2009cadiaplayer, browne2012survey}

%% file: Method.tex
\section{Approach}

In this section, we first derive our objective and then explain how we adapt MCTS to solve it efficiently.

\subsection{Formalization}
\label{sec:Formalization}

\newcommand{\given}{\;\lvert\;}
\newcommand{\IOU}{\text{IoU}}

Given a set $\calI = \{(I_i, D_i)\}_{i=1}^{N_V}$ of $N_V$ registered RGB images and depth maps of a 3D scene, we want to find 3D models and their poses for the objects and walls that constitute the 3D scene. This can be done by looking for a set of objects and layout elements from a pool of proposals, $\hat{\calO}$ that maximizes the posterior given the observations in $\calI$:
\begin{align}
  \hat{\calO} = \arg\max_\calO P( \calO \;\lvert\; \calI ) 
  = \arg\max_\calO \log P( \calO \;\lvert\; \calI ) 
  \> .
\label{eq:posterior}
\end{align}
The set of object proposals contains potential 3D model candidates for each object in the scene, along with its corresponding pose. The same 3D model for an object but under two different poses constitutes two proposals. The set of layout proposals models potential layout candidates as planar 3D polygons. More details about the proposal generation is provided later in Section~\ref{sec:gen_prop}.
% \shreyas{The proposal pool $\calO$ contains potential 3D model candidates for each object in the scene along with its corresponding pose. A 3D model for an object with two different poses constitute two proposals. The layout proposals \shreyasrmk{TODO:Sinisa}. More details about the proposal generation is provided later in Section~\ref{sec:gen_prop}} 

Using the images rather than only the point cloud is important, as shown in~\cite{Qi2020ImVoteNetB3} for example, as many parts of a scanned scene can be missing from the point cloud, when the RGB-D camera did not return depth values for them~(this happens for dark and reflective materials, for example). 
%We represent the objects in $\calO$ by a 3D model and a 6D pose and the layout elements by a 3D polygon.
Assuming the $I_i$ and $D_i$ are independent, $\log P( \calO \given \calI )$ is proportional to:
\begin{align}
\label{eq:score}
    %% & \log P( \calO \given \calI )\nonumber\\
%% \propto &
\sum_i \big( \log P( I_i \given \calO ) + \log P( D_i \given \calO ) \big)
  + \log P( \calO ) \> .
\end{align}

%% Using standard derivations and assumptions, we have:
%% %
%% \begin{align}
%% \label{eq:score}
%%     & \log P( \calO \given \calI )\nonumber\\
%%   \propto &\log P( \calI \given \calO ) P( \calO ) = \log \prod_i P( I_i \given \calO ) P( D_i \given \calO ) P( \calO ) \nonumber\\
%%   = & \sum_i \big( \log P( I_i \given \calO ) + \log P( D_i \given \calO ) \big)
%%   + \log P( \calO ) \> .
%% \end{align}

%%%%%%%%%%%%%%%%%%%%%%%%%%%%%%%%%%%%%%%%%%%%%%%%%%%%%%%%%%%%%%%%%%%%%%%%%%%%%%%%

$P( I_i \given \calO)$ and $P( D_i \given \calO)$ are the likelihoods of our observations.  To evaluate them, we compare $I_i$ and $D_i$ with (non-realistic) renderings of the objects and layout elements in $\calO$ from the same camera poses as the $I_i$ and $D_i$. For $P( I_i \given \calO)$, we render the objects and layout elements in $\calO$ using their class indices in place of colors and compare the result with a semantic segmentation of image $I_i$.  To evaluate $P( D_i \given \calO)$, we render a depth map for the objects and layout elements in $\calO$ and compare it with $D_i$. More formally, we model $\log P( I_i \given \calO ) + \log P( D_i \given \calO )$ by:
\begin{equation}
s_i( \calO ) = \lambda_I \sum_c S_i(c)\cdot S_i^R(c) - \lambda_D  | D_i - D_i^R | \> ,
\label{eq:si}
\end{equation}
up to some additive constant that does not change the optimization problem in Eq.~\eqref{eq:posterior}.  The $S_i(c)$ are segmentation confidence maps for classes $c\in\{\text{wall}, \text{floor}, \text{chair}, \text{table}, \text{sofa}, \text{bed}\}$ obtained by semantic segmentation of $I_i$ (we use MSEG~\cite{lambert2020mseg} for this); the $S_i^R(c)$ are rendered segmentation maps (\ie a pixel in $S_i^R(c)$ has value 1 if lying on an object or layout element of class $c$, 0 otherwise).  $D_i^R$ is the rendered depth map of the objects and layout elements in $\calO$.

Given a set $\calO$, $s_i(\calO)$ can be computed efficiently by pre-rendering a segmentation map and a depth map for each proposal independently: $D_i^R$ can be constructed by taking for each pixel the minimal depth over the pre-rendered depth maps for the proposals in $\calO$. $S_i^R(c)$ can be constructed similarly using both the pre-rendered segmentation and depth maps.

%% The $S_i(c)$ can of course be pre-computed as well.

\begin{figure}
\begin{center}
\scalebox{1}{
\begin{tabular}{cccc}
\multirow{-4}{*}{(a)} & \includegraphics[width=0.3\linewidth]{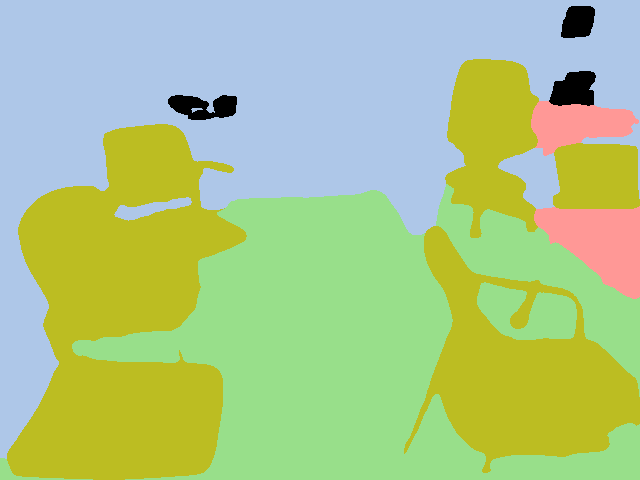} & \multirow{-4}{*}{(b)} & \includegraphics[width=0.3\linewidth]{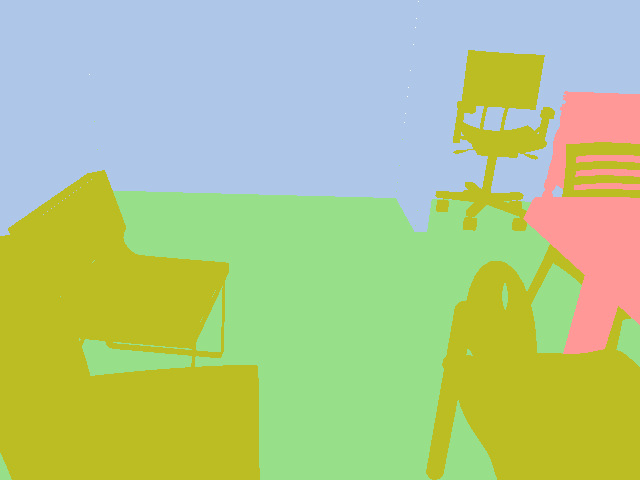} \\
\multirow{-4}{*}{(c)} & \includegraphics[width=0.3\linewidth]{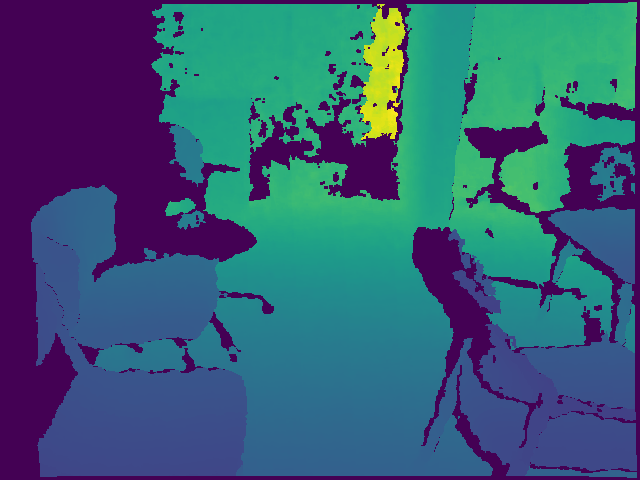} & \multirow{-4}{*}{(d)} &  \includegraphics[width=0.3\linewidth]{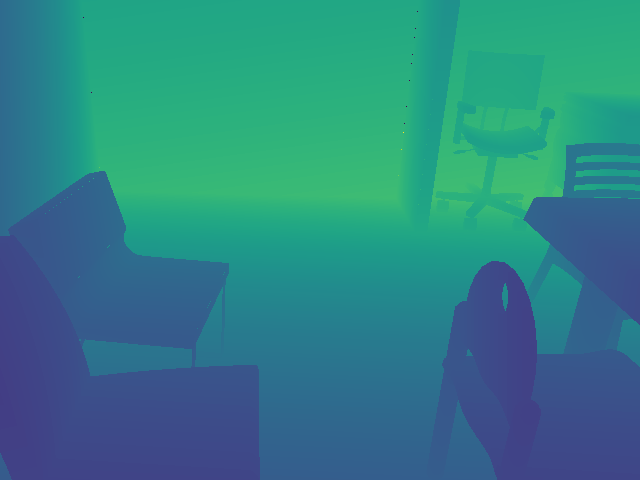} \\
\end{tabular}}
\end{center}
\vspace{-0.5cm}
\caption{\label{fig:SandR} Examples for (a) $S_i$, (b) $S_i^R$, (c) $D_i$, (d) $D_i^R$.}
\vspace{-0.5cm}
\end{figure}

Fig.~\ref{fig:SandR} shows an example of $S_i$, $S_i^R$, $D_i$, and $D_i^R$.  Note that our approach considers all the objects together and takes naturally into account the occlusions that may occur between them, which is one of the advantages of analysis-by-synthesis approaches.  More sophisticated ways to evaluate the observations likelihoods could be used, but this simple method already yields very good results.

%%%%%%%%%%%%%%%%%%%%%%%%%%%%%%%%%%%%%%%%%%%%%%%%%%%%%%%%%%%%%%%%%%%%%%%%%%%%%%%%

%% This term could be used for example to encourage realistic scenes.  For simplicity, we use it to avoid physically infeasible solutions. In such cases, $P(\calO)$ should be equal to 0.

$P(\calO)$ in Eq.~\eqref{eq:score} is a prior term on the set $\calO$.  We currently use it to prevent physically impossible solutions only.  In practice, the proposals are not perfectly localised and we tolerate some intersections.  When the Intersection-Over-Union between two objects is smaller than a threshold, we tolerate the intersection but still penalize it. More formally, in this case, we model $\log P( \calO )$ by
\begin{align}
  s^p( \calO ) = - \lambda_P \sum_{O, O'\in\calO,O\neq O'} \IOU(O, O') \> 
\end{align}
up to  some additive constant. $\IOU$ is the intersection-over-Union between the 3D models for objects $O_j$ and $O_k$.  In practice, we compute it using a voxel representation of the 3D models.  When the Intersection-over-Union between two object proposals is above a threshold, we take $P(\calO) = 0$, \ie the two proposals are incompatible. In practice, we use a threshold of 0.3. We consider two special cases where this is not true: chair-table and sofa-table intersections. In these cases, we first identify the horizontal surface on which the intersection occurs (\eg surface of the table, seat of the sofa) and determine the amount of intersection by calculating the distance of the intersecting point to nearest edge of the horizontal surface. The amount of intersection is normalized by the dimension of the horizontal surface and a ratio more than 0.3 is considered incompatible.

Similarly, when two layout proposals intersect or when a layout proposal and an object proposal intersect, we take also $P(\calO) = 0$.  In contrast to object proposals where small intersections are still tolerated, we do not tolerate any intersections for the layout proposals as their locations tend to be predicted more accurately.

%% \vincent{Moreover, our method retrieves closed (possibly non-convex) polyhedra for the layouts, 

%% \vincentrmk{We need to formalize this better: }
%% \sinisa{ To enforce structured layouts, we add $0.1$ to the score if layout components in $\calO$ build a ring.  }  \sinisarmk{Can we at least say we penalize solutions with infeasible layout shapes, \eg extremely small distances between disconnected corners, etc.?  This is instead of visibility term for layouts.}

As discussed in the introduction, to find a set $\hat{\calO}$ that maximizes Eq.~\eqref{eq:posterior}, we build a pool $\calO_\pool$ of proposals, and select $\hat{\calO}$ as the subset of $\calO_\pool$ that maximizes the global score $S(\calO) = \sum_i s_i(\calO) + s^P(\calO)$. 
We empirically set $\lambda_I = \lambda_D$ = 1 and $\lambda_P = 2.5$ in our experiments to balance the three terms in Eq.~\eqref{eq:score}.

\subsection{Monte Carlo Scene Search}
\label{sec:mcss}

We now explain how we adapted MCTS to perform an efficient optimization of the problem in Eq.~\eqref{eq:score}.  We call this variant ``Monte Carlo Scene Search''~(MCSS).

\subsubsection{Tree Structure}

In the case of standard MCTS, the search tree follows directly from the rules of the game.  We define the search tree explored by MCSS to adapt to the scene understanding problem and to allow for an efficient exploration as follows.

{\bf Proposal fitness.}  Each proposal $\calP$ is assigned a \textit{fitness} value obtained by evaluating $s_i$ in Eq.~\eqref{eq:si} only over the pixel locations where the proposal reprojects.  Note that this fitness is associated with a proposal and not a node. This fitness will guide both the definition and the exploration of the search tree during the simulations.

Except for the root node, a node $\calN$ in the scene tree is associated with a proposal $\calP(\calN)$ from the pool $\calO_\pool$.  Each path from the root node to a leaf node thus corresponds to a set of proposals $\calO$ that is a potential solution to Eq.~\eqref{eq:posterior}.  We define the tree so that no path can correspond to an impossible solution \ie to set $\calO$ with $P(\calO)=0$.  This simplifies the search space to the set of possible solutions only.  We also found that considering first proposals that are close spatially to proposals in a current path significantly speeds up the search, and we also organize the tree by spatial neighbourhood. The child nodes of the root node are made of a node containing the proposal $O$ with the highest fitness among all proposals, and a node for each proposal that is incompatible with $O$.  The child nodes of every other node $\calN$ contain the closest proposal $O$ to the proposal in $\calN$, and the proposals $O'$ incompatible with $O$, under the constraint that $O$ and proposals $O'$ are compatible with all the proposals in $\calN$ and its ancestors.

Two layout proposals are considered incompatible if they intersect and are not spatial neighbours. They are spatial neighbors if they share an edge and are not on the same 3D plane. Therefore, if $\calP(\calN)$ is a layout proposal, the children nodes are always layout components that are connected by an edge to $\calP(\calN)$. By doing so, we enforce that each path in the tree enforces structured layouts, \ie the layout components are connected. Note that this strategy will miss disconnected layout structures such as pillars in the middle of a room but works well on ScanNet.

In the case of objects, the spatial distance between two object proposals is computed by taking the Euclidean distance between the centers of the 3D bounding boxes. The incompatibility between two object proposals is determined as explained in Section~\ref{sec:Formalization}. Since all the object proposals in the children of a node may be all incorrect, we add a special node that does not contain a proposal to avoid having to select an incorrect proposal. The children nodes of the special node are based on the proximity to its parent node excluding the proposals in its sibling nodes.

As mentioned in the introduction, we first run MCSS on the layout component proposals only to select the correct layout components first. Then, we run MCSS on the object proposals, with the selected layout components in $\calO$. The selection of the object proposals therefore benefits from the recovered layout.

%% Our scene tree is built by connecting a ``layout tree'' and ``object trees''. Every leaf node of the layout tree is the root node of an object tree. Next, we discuss the specifics of the layout and object trees. In practice, we first search the layout tree only for a given amount of iterations. After that, the layout tree is frozen, and the best path in the tree is used as the root node of the object tree.

\subsubsection{Local node scores}

% Note that this consistent with the observation of \cite{schadd2008single}, who noticed that in the interest of computation time, exploiting local maximums can be more beneficial than exploiting current global maximums.

Usually with MCTS, $\ScoreSum$ in the UCB criterion given in Eq.~\eqref{eq:UCB} and stored in each node is taken as the sum of the game final scores obtained after visiting the node.  We noticed during our experiments that exploration is more efficient if $\ScoreSum$ focuses more on views where the proposal in the node is visible. Thus, in MCSS, after a simulation returns $\calO$, the score $s$ is added to $\ScoreSum$ of a node containing a proposal $O$. $s$ is a local score calculated as follows to focus on $O$:
\begin{align}
s = \frac{1}{\sum_i w_i(O)} \sum_i w_i(O) s_i(\calO) + \lambda_p s^P(O, \calO)  \> ,
\label{eq:localscore}
\end{align}
where $w_i(O) = 1$ if $O$ is visible in view $i$ and 0 otherwise, and
\begin{align}
s^p( O, \calO ) = - \sum_{O'\in\calO,O\neq O'} \IOU(O, O') \> .
\end{align}

\subsubsection{Running simulations}

While running the simulations, instead of randomly picking the nodes, we use a ``roulette wheel selection'' based on their proposals: the probability for picking a node is directly proportional to the \textit{fitness} of the proposal it contains.

%% } \sinisarmk{No. Simulation does not depend on selection criterion. There was an additional paragraph stating that instead of randomly selecting proposals in simulation, we assign probabilities for picking the proposals based on fitness (this is usually called roulette wheel selection). That means, we should remove third term in selection criterion and updating proposal fitnesses section. Instead we should explain how simulation is adapted.}

%% \subsubsection{Updating proposal fitnesses}

%% Every time including a component $O$ in a candidate solution obtains the new maximum score for any of the views where $O$ is visible, we increase its fitness by $0.1$. If the average score for these views is lower than $0.5$ of the average of the best scores achieved for these views so far, we decrease the fitness by $0.01$. By doing so, we are able to propagate information from all the nodes containing this proposal: Visiting a node containing some proposal will therefore influence the selection of nodes containing the same proposal.

\subsubsection{MCSS output}
Besides the tree definition and the local score given in Eq.~\eqref{eq:localscore} used in the \textsc{Select} criterion, MCSS runs as MCTS to return the best set $\calO$ of proposals found by the simulations according to the final score $S(\calO) = \sum_i s_i(\calO) + s^P(\calO)$. In practice, we perform 20,000 iterations of MCSS.

%% $\log(\calO \given \calI)$.

\subsection{Generating Proposals}
\label{sec:gen_prop}

% \vincent{ We resort here on off-the-shelf techniques.  For the object proposals, we trained VoteNet~\cite{qi2019deep} on simple synthetic point clouds to generate 3D bounding boxes with their predicted classes.  Using simple heuristics, we also generate additional 3D bounding boxes by splitting and merging the detections from VoteNet.  We also obtain a semantic segmentation of the point cloud by backprojecting the image segmentations from MSEG~\cite{lambert2020mseg} using majority voting to merge the segmentations from multiple views.  This is useful to remove the points inside the bounding boxes that do not belong to the predicted class.  We trained a network based on PointNet++~\cite{qi2017pointnetplusplus} on the same synthetic data to predict an embedding for a CAD model from ShapeNet~\cite{shapenet} and a 6D pose+scale from samplings of the remaining points.  Different samplings result in slightly different embeddings and we generate a proposal with each of the corresponding CAD models.  We refine the pose and scale estimates by performing a small grid search around the predicted values using the Chamfer distance between the CAD model and the point cloud.}

% Such a textureless, point-cloud only dataset is highly generalizable as we show later in the evaluations that networks trained to generate proposals on this dataset work well on ScanNet scenes and in-the-wild RGBD scans which we capture ourselves. 

We resort here on off-the-shelf techniques.  For the object proposals, we first create a set of synthetic point clouds  using ShapeNet~\cite{shapenet} CAD models and the ScanNet dataset~\cite{dai2017scannet}~(we provide more details in the suppl. mat.). We train VoteNet~\cite{qi2019deep} on this dataset to generate 3D bounding boxes with their predicted classes. Note that we do not need VoteNet to work very well as we will prune the false positives anyway, which makes the approach generalizable. 
 Using simple heuristics, we create additional 3D bounding boxes by splitting and merging the detections from VoteNet, which we found useful to deal with cluttered scenes. We also train MinkowskiNet~\cite{choy20194d} on the same synthetic dataset which we use to remove the points inside the bounding boxes that do not belong to the Votenet predicted class.  We then trained a network based on PointNet++~\cite{qi2017pointnetplusplus} on the same synthetic data to predict an embedding for a CAD model from ShapeNet~\cite{shapenet} and a 6D pose+scale from samplings of the remaining points.  Different samplings result in slightly different embeddings and we generate a proposal with each of the corresponding CAD models.  We refine the pose and scale estimates by performing a small grid search around the predicted values using the Chamfer distance between the CAD model and the point cloud.

For the layout component proposals, we use the semantic segmentation by  MinkowskiNet to extract the 3D points on the layout from the point cloud and rely on a simple RANSAC procedure to fit 3D planes.  Like previous works~\cite{murali2017indoor, nan2017polyfit, ZouGLH19, stekovic2020general}, we compute the intersections between these planes to obtain 3D polygons, which we use as layout proposals. We also include the planes of the point cloud's 3D bounding box faces to handle incomplete scans: for example, long corridors are never scanned completely in ScanNet.

%% file: Evaluations.tex
\section{Evaluation}
\label{sec:evaluations}

We present here the evaluation of our method. We also provide an ablation study to show the importance of our modifications to MCTS and of the use of the retrieved layouts when retrieving the objects. 

Fig.~\ref{fig:custom_scan} shows the output of our method on a custom scan, and more qualitative results are provided in the suppl.~mat.

% We finally provide additional results on our own scans.

\newcommand{\niceresultwidth}{0.18\linewidth}
\newcommand{\niceresult}[1]{
\multirow{-5}{*}{(a)} &
\includegraphics[trim=0 10 0 10,clip, width=\niceresultwidth]{figures/qualitative/#1_scene.png} & \multirow{-5}{*}{(b)} &
\includegraphics[trim=0 10 0 10,clip,width=\niceresultwidth]{figures/qualitative/#1_baseline.png} & \multirow{-5}{*}{(c)} &
\includegraphics[trim=0 10 0 10,clip,width=\niceresultwidth]{figures/qualitative/#1_our.png} & \multirow{-5}{*}{(d)} &
\includegraphics[trim=0 10 0 10,clip,width=\niceresultwidth]{figures/qualitative/#1_gt.png} \\
}
\begin{figure*}
    \centering
    \scalebox{.99}{
    \begin{tabular}{cccccccc}
        \niceresult{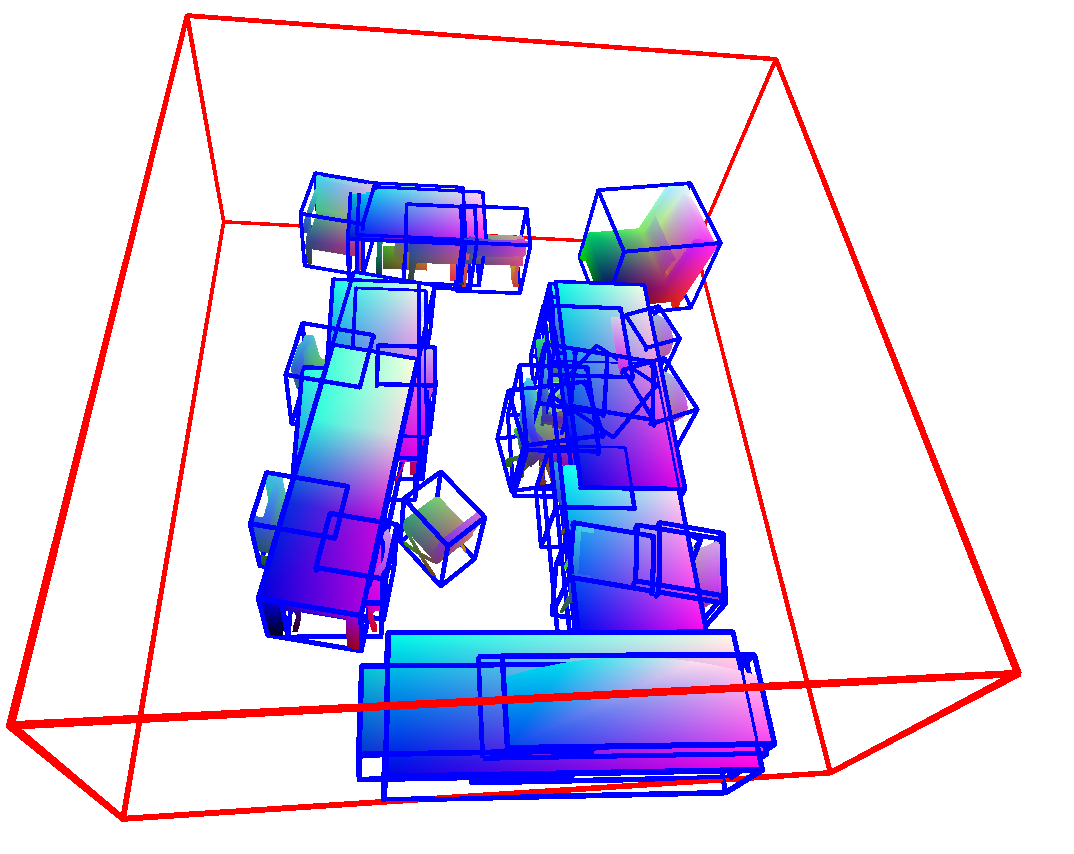}
        %  \multicolumn{4}{c}{    \includegraphics[width=0.9\linewidth]{figures/qualitative/qual1.png}} \\ 
        % (a) & (b) & (c) & (d) \\
    \end{tabular}}
    \vspace{-0.4cm}
    \caption{(a) An RGB-D scan from the ScanNet dataset~\cite{dai2017scannet}. (b) Output of the VoteNet-based baseline method for the objects, together with the layout annotations from \cite{avetisyan2020scenecad}. Many  objects retrieved by the baseline method are incorrect; the layout annotations lack some details. (c) Objects and layout prediction by our MCSS method. Our predicted layout has much more details than the manual annotations. (d) Objects annotations from~\cite{Avetisyan_2019_CVPR} together with our manual layout annotations. The supp.~mat.~provides more visualizations.}
    \label{fig:my_label}
    \label{fig:qual_results}
    \vspace{-0.3cm}
\end{figure*}

\newcommand{\customniceresultwidth}{0.45\linewidth}
\newcommand{\customniceresult}[1]{
\includegraphics[width=\customniceresultwidth]{figures/qualitative/#1_scene.png} &
\includegraphics[width=\customniceresultwidth]{figures/qualitative/#1_our.png}\\
}

\begin{figure}
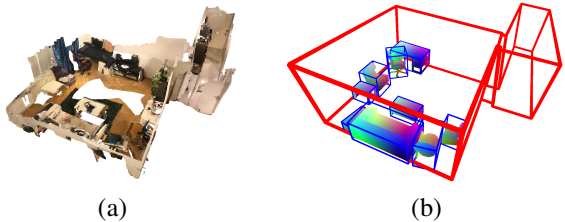

    \centering
    \begin{tabular}{cc}
    \customniceresult{shreyas_apt_007}
         (a) & (b) \\
    \end{tabular}
    \vspace{-0.2cm}
    \caption{{\bf Generalization to other datasets. } (a) We captured an RGB-D scan of an apartment with a hallway and a living space, and many furniture. (b) Objects and layout found by our MCSS method. More results are provided in the suppl.~mat.}
    \vspace{-0.5cm}
    \label{fig:custom_scan}
\end{figure}

% On the SceneCAD~\cite{avetisyan2020scenecad} dataset, 

\subsection{Layouts}

We first evaluate the ability of MCSS to recover general layouts on validation scenes from the SceneCAD dataset~\cite{Avetisyan_2019_CVPR}  that provides layout annotations for noisy RGBD scans from the ScanNet dataset~\cite{dai2017scannet}. MCSS outperforms the SceneCAD method by a quite substantial margin on the corner recall metric, with 84.8\% compared to 71\%. However, as shown in Fig.~\ref{fig:qual_results}(b), the SceneCAD annotations lack details, which hurts the  performance of our method on other metrics as it recovers details not in the manual annotations.

Hence, we relabelled the same set of scenes from the SceneCAD dataset with more details. As proposed in the SceneCAD paper, a predicted corner is considered to be matching to the ground truth corner if it is within $40cm$ radius. We further adjust this criterion: if multiple predicted corners are within this radius, a single corner that is closest to the ground truth is taken and a predicted corner can be assigned to only one ground truth corner. We also compute the polygons' Intersection-Over-Union~(IOU) metric from \cite{stekovic2020general} after projecting the retrieved polygons to their ground truth polygons. Table~\ref{tab:our_layout_comp} compares the layouts retrieved by our approach to the SceneCAD  annotations. These annotations  obtain very high corner precision, as most of the annotated corners are indeed correct, but low corners recall and polygon IOU because of the missing details. By contrast, our method recovers most corners which results in high recall without generating wrong ones, as is visible from the high precision. Our approach does well to recover general room structure as shown by the polygon IOU value. We show in Fig.~\ref{fig:qual_results},~\ref{fig:custom_scan} and suppl. mat. that our method successfully recovers a variety of layout configurations. Most errors come from the fact that components might be completely invisible in the scene in all of the views as our proposal generation is not intended for this special case. 

% Also, designing score functions that are more specific to the layout estimation task can help to alleviate problems in cases where semantic segmentation and depth terms are not reliable.

% On top of that, SceneCAD method has seen the same type of annotations in the training set while our method does not rely on such strong bias. On other metrics our method shows lower numbers which do not reflect performance of our method. The reason is that our method is able to recover much higher amount of detail in the scene, even when compared to the manual annotations from the SceneCAD dataset. \sinisarmk{Can we say it like this?}

\begin{table}[]
    \centering
    \scalebox{.8}
	     {
    \begin{tabular}{l|c c c|c c c }
    & \multicolumn{3}{c}{All Scenes} & \multicolumn{3}{c}{Non-Cuboid Scenes}  \\
    & Prec & Rec & IOU &  Prec & Rec & IOU \\
    \hline
    SceneCAD GT & \textbf{91.2} & 80.4 & 75.0 & \textbf{90.8} & 73.3 & 66.1 \\
     MCSS (Ours) &  85.5 & \textbf{86.1} & \textbf{75.8} & 83.5 & \textbf{80.4} & \textbf{70.4} \\
    \end{tabular}}
    \caption{Comparison between \emph{manual} SceneCAD layout annotations and layouts \emph{retrieved} by our method, on our more detailed layout annotations.}
    \label{tab:our_layout_comp}
    \vspace{-0.7cm}
\end{table}

% \begin{table*}[]
%     \centering
%     \begin{tabular}{l||c c c c c | c c c c c||}
%     & \multicolumn{4}{c}{All Validation Scenes} && \multicolumn{4}{c}{Non-Cuboid Scenes} & \\
%     \hline
%     & \multicolumn{2}{c}{Corners (\%)} & \multicolumn{2}{c}{Poly.  (\%)} && \multicolumn{2}{c}{Corners (\%)} & \multicolumn{2}{c}{Corners (\%)} & \\ 
%     & Precision & Recall & \multicolumn{2}{c}{IOU} &&  Precision & Recall & \multicolumn{2}{c}{IOU} & \\
%         \hline

%         \shortstack{SceneCAD~\cite{avetisyan2020scenecad} (g. t.)} & \textbf{92.1} & 79 & \multicolumn{2}{c}{72} && \textbf{92.2} & 70.8 & \multicolumn{2}{c}{66.3} \\
%         Our &  86 & \textbf{89.2} & \multicolumn{2}{c}{\textbf{79.4}}&& 84.4 & \textbf{84.6} & \multicolumn{2}{c}{\textbf{72.4}} &
%     \end{tabular}
%     \caption{Comparison between SceneCAD layout annotations and layouts retrieved by our method, on our more detailed layout annotations.}
%     \label{tab:our_layout_comp}
% \end{table*}

\subsection{Objects}
\label{sec:eval_objs}
We evaluate our method on the subset of scenes from both the test set and validation set of Scan2CAD~\cite{Avetisyan_2019_CVPR}. We consider 95 scenes in the test set and 126 unique scenes in the validation which contains at least one object from the \textit{chair, sofa, table, bed} categories. A complete list of the scenes used in our evaluations is provided in the suppl. mat.

% We consider 95 of 97 scenes from the test set. The two scenes are not considered due to inconsistency in annotations and presence of multiple floor planes, a scenario that we do not consider in our Object tree. 

% \vincentrmk{why do you retrieve more bounding boxes when generating the proposals?}\shreyasrmk{this is addressed in object proposals} 

We first consider a baseline  which uses Votenet~\cite{qi2019deep} for object detection and retrieves a CAD model and its pose for each 3D bounding box using the same network used for our proposals. The performance of this baseline will show the impact of not using multiple proposals for both object detection and model retrieval.

\begin{figure}
    \centering
    \scalebox{1.}{
    \begin{tabular}{cc}
        \includegraphics[trim=20 20 20 20,clip,width=0.35\linewidth]{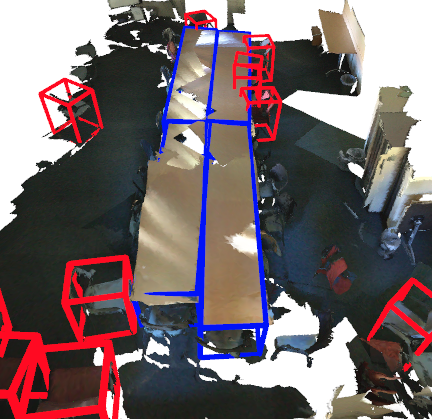} & \includegraphics[trim=20 20 20 20,clip,width=0.35\linewidth]{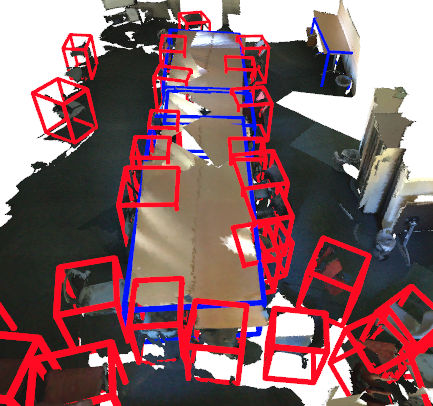}  \\ 
         (a) Manual Annotations & (b) MCSS (ours)\\
    \end{tabular}}
      \vspace{-0.2cm}
    \caption{{\bf Manual annotation of incomplete point clouds is difficult} and groups of tables are often annotated incorrectly in the Scan2CAD dataset, creating a dataset bias. For example, in (a), there should be 8 tables instead of 4 in the annotations. This hurts our performance for the \textit{table} category, though we achieve plausible solutions (b). Note that we also often retrieve more  objects than in the annotations. }
  \label{fig:table_anno}
  \vspace{-0.5cm}
\end{figure}

We use the accuracy metric defined in \cite{Avetisyan_2019_CVPR} for evaluations on the test set and compare with three methods~(  Scan2CAD~\cite{Avetisyan_2019_CVPR}, E2E~\cite{Avetisyan2019EndtoEndCM}, and SceneCAD~\cite{avetisyan2020scenecad}) in Table~\ref{tab:testset_obj}. While our method is trained only on simple synthetic data, it still outperforms Scan2CAD and E2E on the \textit{chair} and \textit{sofa} categories. The loower performance on the \textit{table} category is due to inconsistent manual annotations: Instance level annotation of a group of tables from an incomplete point cloud is challenging and this results in inconsistent grouping of \textit{tables} as shown in Fig.~\ref{fig:table_anno}. Although we achieve plausible solutions in these scenarios, it is difficult to obtain similar instance-level detection as the manual annotations.  Moreover, SceneCAD learns to exploit object-object and object-layout support relationships, which significantly improves the performance. Our approach does not exploit such constraints yet, but they could be integrated in the objective function's prior term in future work for benefits.

Table~\ref{tab:valset_obj_cd} compares the Chamfer distance between the objects we retrieve and the manually annotated point cloud of the object on the  validation set of ScanNet. 
This metric captures the accuracy of the retrieved CAD models. The models we retrieve for \textit{chair} and \textit{sofa}  are very similar to the models chosen for the manual annotations as the Chamfer distances have the same order of magnitude.

Table~\ref{tab:valset_obj_iou} reports the precision and recall for the oriented 3D bounding boxes for the pool of object proposals, for the set of proposals selected by MCSS, and for the baseline. MCSS improves the precision and recall from the baseline  in all 4 object categories. The recall remains similar while the precision improves significantly. This proves that our method efficiently rejects all incorrect proposals. Our qualitative results in Fig.~\ref{fig:qual_results} and \ref{fig:table_anno} show the efficacy of MCSS in rejecting many incorrect proposals compared to the baseline method while also retaining the correct CAD models that are similar to ground truth. We even retrieve objects missing from the annotations.

\begin{table}
    \centering
        \scalebox{.7} 
	     {
    \begin{tabular}{l| c | c c c c c c c c}
    & \centercell{0.6cm}{IOU Th.} & \multicolumn{2}{c}{Chair} & \multicolumn{2}{c}{Sofa} & \multicolumn{2}{c}{Table} & \multicolumn{2}{c}{Bed} \\
  & & Prec & Rec & Prec & Rec & Prec & Rec & Prec & Rec \\
    \hline
    \centercell{1.2cm}{All proposals} & 0.50 & 0.06 & 0.92 & 0.05 & 0.93 & 0.05 & 0.68 & 0.16 & 0.93 \\
    & 0.75 & 0.04 & 0.59 & 0.04 & 0.56 & 0.03 & 0.46 & 0.08 & 0.48 \\   [0.5ex]
    
    \centercell{1.2cm}{Baseline} & 0.50 & 0.70 & 0.85 & 0.77 & 0.80 & 0.66 & 0.56 & 0.74 & 0.74 \\
    & 0.75 & 0.19 & 0.29 & 0.31 & 0.39 & 0.24 & 0.30 & 0.30 & 0.41 \\[0.5ex]
    \centercell{1.2cm}{MCSS (Ours)} & 0.50 & 0.75 & 0.87 & 0.79 & 0.93 & 0.65 & 0.59 & 0.86 & 0.86 \\
    & 0.75 & 0.27 & 0.32 & 0.42 & 0.42 & 0.34 & 0.30 & 0.41 & 0.44 
    
    % \hline
    % %
    % & 0.75 & 0.04 & 0.59 & 0.04 & 0.56 & 0.03 & 0.46 & 0.08 & 0.48 \\ 
    % & 0.75 & 0.19 & 0.29 & 0.31 & 0.39 & 0.24 & 0.30 & 0.30 & 0.41 \\ 
    % & 0.75 & 0.27 & 0.32 & 0.42 & 0.42 & 0.34 & 0.30 & 0.41 & 0.44 \\
    
    % % \parbox[t]{3.5cm}{Votenet + Model Retr. \newline (baseline)} & 0.70 & 0.85 & 0.77 & 0.80 & 0.66 & 0.56 & 0.74 & 0.74 \\
    % MCSS (Ours) & 0.75/0.27 & 0.87/0.32 & 0.79/0.42 & 0.93/0.42 & 0.65/0.34 & 0.59/0.30 & 0.86/0.41 & 0.86/0.44 
    \end{tabular}}
    \caption{{\bf Evaluation of object model retrieval and alignment} with bounding box IOU thresholds 0.5 and 0.75. The recall for our method is similar to the recall with all proposals while precision is better than the baseline method. Our method efficiently rejects all the incorrect proposals.}
    \vspace{-0.3cm}
    \label{tab:valset_obj_iou}
\end{table}

\begin{table}
\centering
\scalebox{0.75}{
\begin{tabular}{c |c| c c c} 
 Method & Obj-Obj Support & Chair & Sofa & Table \\
 \hline
 Baseline  & No & 42.02 & 27.70 & 18.52 \\
 Scan2CAD \cite{Avetisyan_2019_CVPR} & No & 44.26 & 30.66 & 30.11 \\ 
 E2E \cite{Avetisyan2019EndtoEndCM} & No & 73.04 & 76.92 & \textbf{48.15} \\
 SceneCAD \cite{avetisyan2020scenecad} & Yes & 81.26 & 82.86 & 45.60 \\
 MCSS (Ours) & No & \textbf{74.32} & \textbf{78.70} & 24.28 \\
%  \multicolumn{5}{c}{} \\
%  \hline
 \end{tabular}}
%  \vspace{-0.5cm}
\caption{{\bf Comparison of object alignment on the Scan2CAD benchmark.} The metrics for \textit{bed} alone are not provided by the benchmark and hence not shown. SceneCAD uses inter-object support relations to improve their results from E2E. We do not have access to these relationships and hence mostly compare with E2E and Scan2CAD. The lower accuracy for \textit{table} seems to be due to the dataset bias discussed in Fig.~\ref{fig:table_anno}.}
\label{tab:testset_obj}
\vspace{-0.2cm}
\end{table}

\begin{table}
\centering
\scalebox{0.8}{
\begin{tabular}{c | c c c c} 
 Method & Chair & Sofa & Table & Bed\\ [0.5ex] 
 \hline
  Baseline & 2.6 & 11.0 & 14.2 & 26.3 \\ 
%  Ours (before pose refine) & 2.3 & 7.7 & 12.3 & 18.6 \\
%  \hline
 MCSS (Ours)  & 1.8 & 7.4 & 12.8 & 16.2 \\
 Manual annotations~\cite{Avetisyan_2019_CVPR} & 2.0 & 5.2 & 5.5 & 9.4 \\
\end{tabular}}
\vspace{-0.3cm}
\caption{{\bf Comparison of one-way Chamfer distance (in mm) between scan points and retrieved models on the validation set of Scan2CAD.} Our retrieved models are close to manual annotations for \textit{chair} and \textit{sofa} even though we use only synthetic point clouds for model retrieval.}
\label{tab:valset_obj_cd}
\vspace{-0.5cm}
\end{table}

\subsection{Ablation Study}
\begin{figure}[h]%{\linewidth}
    \centering
    \begin{tabular}{cc}
    \multirow{-8}{*}{} &
    \includegraphics[trim=0 0 0 0,clip, width=.8\linewidth,height=0.35\linewidth]{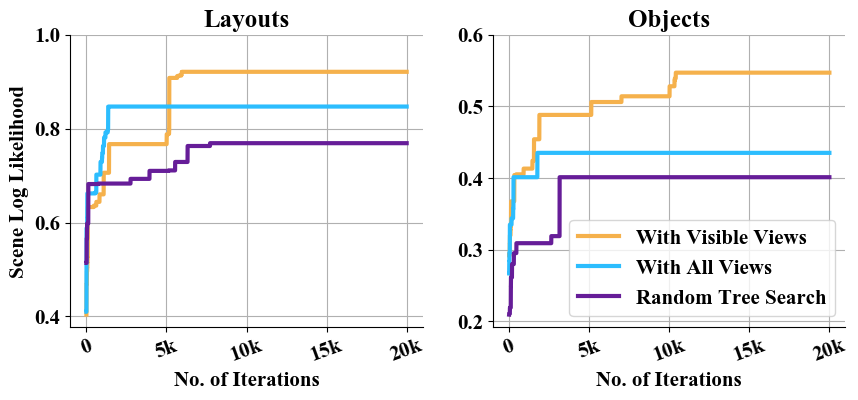} \\
    
    & 
    \scalebox{.7}
	     {
    \begin{tabular}{c | c c } 
         Search Method & Node Score & Simulation \\% [0.5ex] 
         \hline
         \centering With Visible Views & Eq.~\eqref{eq:localscore} & Roulette Wheel \\ 
         With All Views &$\sum_i s_i(\mathcal{O}) + s^P(\mathcal{O})$ &  Roulette Wheel \\
         \multirow{1}{*}{\parbox{2.4cm}{\centering Random Search}} & \multirow{1}{*}{\parbox{2.4cm}{\centering 0}} & Uniform Probability \\
        %  &&\\
    \end{tabular}} \\
    \end{tabular}
    \vspace{-0.2cm}
    %   \captionof{table}{}
    \caption{Best  score $S(O) = \big(\sum_i s_i(\calO) + s^P(\calO)\big)$ found so far for layout and objects over MCSS iterations. Using the local score given in Eq.~\eqref{eq:localscore} results in much faster and better convergence. }
    \label{fig:local_global}
    \vspace{-0.8cm}
\end{figure}

\vspace{-0.1cm}
\paragraph{Importance of local score (Eq.~\ref{eq:localscore}).} 
In Fig.~\ref{fig:local_global},
we plot the best score $S(\calO)$
 found so far with respect to the MCTS iteration, in the case of a complex scene for layout recovery and object recovery, when using the simulation score  $S(\calO)$ or the local score $s$ given in Eq.~\eqref{eq:localscore} to update the $Q$ of the nodes. We use the selection strategy of Eq.~\eqref{eq:UCB} in both of these scenarios. We also plot the best score for a random tree search. Using the local score speeds up the convergence to a better solution, achieving on an
average 9\% and 15\% higher global scores for layouts and objects, respectively. Compared to random tree search, our method achieves 15\% and 42\% higher  scores for layout and objects, respectively. We consider 12 challenging scenes for this experiment.

\paragraph{Importance of layout  for retrieving objects.} Table~\ref{tab:ablation_obj_layout} shows the effect of using the estimated layout in the terms of Eq.~\eqref{eq:si} while running MCSS on objects. We considered 12 challenging scenes mainly containing chairs and tables for this experiment and use the same precision and recall metrics as in Table~\ref{tab:valset_obj_iou}. Using the layout clearly helps by providing a better evaluation of image and depth likelihoods.

\begin{table}[h]
    \centering
    \scalebox{0.75}{
    \begin{tabular}{l|c c c c}
    & \multicolumn{2}{c}{Chair} & \multicolumn{2}{c}{Table} \\[-0.1cm]
    & Prec & Rec & Prec & Rec \\
    \hline
        Without layout &  0.58 & 0.61 & 0.48 & 0.34 \\
        With layout & 0.65 & 0.84 & 0.66 & 0.58
    \end{tabular}}
    \vspace{-0.3cm}
    \caption{\bf Impact of using the estimated layout when running MCSS for object retrieval.}
    \label{tab:ablation_obj_layout}
    \vspace{-0.5cm}
\end{table}

%  on the IOU metric for 12 challenging scenes.} 

%% file: Supplementary.tex
In this supplementary material:
\begin{itemize}
\item we suggest some possible future directions, 
\item we detail our methods for generating layout and object proposals, and give the pseudocode for MCTS for reference,
\item we provide additional comparisons with existing annotations, the results of our MCSS approach, and a baseline using hill climbing for the optimization of our objective function,
\item we provide more qualitative results on scans outside the ScanNet dataset. 
\end{itemize}

In addition to this document, we provide a \textbf{Supplementary Video} showing the improvement of the solution found by MCSS over time, and  additional qualitative demonstrations.

\section{Future Directions}

While MCSS usually recovers all objects in a scene and complete layouts as we can use low thresholds when generating the proposals without returning false positives, there are still  situations where it is challenging to retrieve the correct object models or layout components, when the point cloud misses too much 3D data. 

There are still many directions in which our current method could be improved. We could generate proposals from the perspective views as well: RGB images often contain useful information that is missing in the point cloud, and we can handle many proposals. Comparing the final solution with the RGB-D data could also be used to detect objects or layout components that are not explained by the solution, and could be integrated as additional proposals in a new run of MCSS.  To improve the 3D poses and models, it would also be interesting to develop a refinement method that improves all the identified objects together.

% As proposals are generated from noisy and incomplete point clouds, in some situations it is still very challenging to retrieve correct object models or layout components. One interesting addition to our proposal generation pipeline would be to  consider data from perspective views when generating proposals. RGB images often contain information that is missing in the point cloud. 

Furthermore, advanced MCTS-based algorithms such as AlphaZero~\cite{AlphaZero} utilize neural networks to evaluate the quality of state-action pairs. Similarly, it should be possible to train a deep network to predict which proposals should be evaluated first. We thus believe that our approach opens new directions to explore.

% neural networks could be trained to estimate how well the proposed solution fits the scene. 

% Further, synthetic scene datasets can potentially be used to learn to enforce relationship between objects, objects and layout components, and with the overall scene context.
% Also nerual network could be trained to evaluate the prior term, \ie proposal-proposal relations.

%%%%%%%%%%%%%%%%%%%%%%%%%

\section{Layout Proposal Generation}

\begin{figure*}
    \centering
    \begin{tabular}[t]{cccc}
         \includegraphics[width=0.23\linewidth]{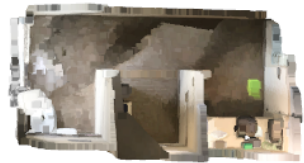} &  
         \includegraphics[width=0.23\linewidth]{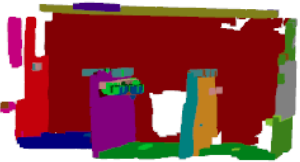} & 
         \includegraphics[width=0.23\linewidth]{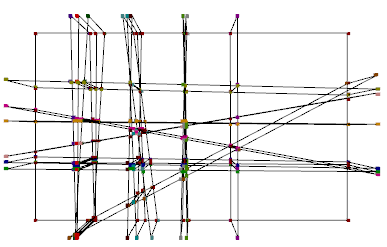} &
         \includegraphics[width=0.23\linewidth]{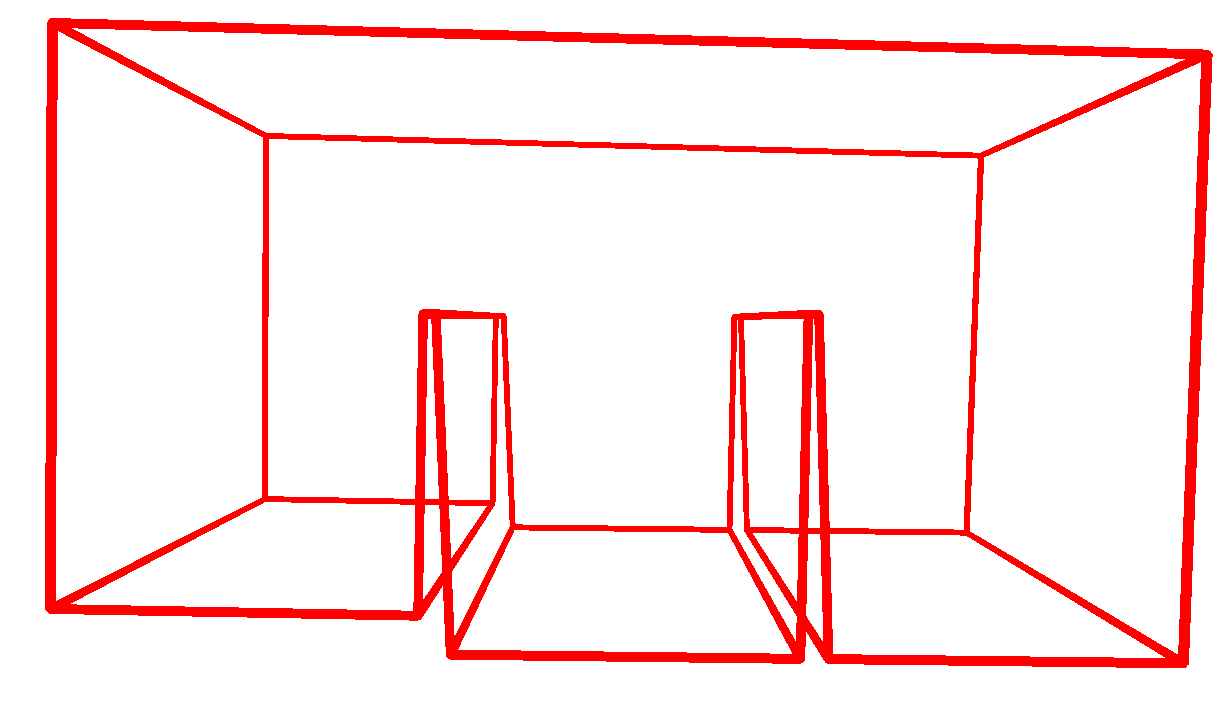} \\
         Input point cloud & Layout planes segmentation & Layout proposals & Reconstructed Layout \\
    \end{tabular}
    \caption{We detect layout planes from the input point cloud using our RANSAC procedure. By intersecting these planes, we obtain a large number of planar polygons which we take as our layout proposals. MCSS selects the optimal subset of proposals that best fits the input scene.}
    \label{fig:lay_prop_gen}
\end{figure*}

Figure~\ref{fig:lay_prop_gen} describes our layout proposal generation. We first detect planes that are likely to correspond to layout components (walls and floors in our experiments). Based on the output from MinkowskiNet~\cite{choy20194d}, we remove from the point cloud the 3D points that do not belong to layout classes, and perform RANSAC plane fitting on the remaining points. We implemented a variant of RANSAC, using 3-point plane fitting that determines inlier-points by their distance and their normals orientation with respect to the sampled plane. We only fit a single floor plane as the SceneCAD dataset~\cite{avetisyan2020scenecad} does not contain any scenes with multiple floor planes.

At each iteration, our RANSAC procedure fits a plane to three points that are randomly sampled from the remaining point cloud. The inliers are defined as a set of points in the point cloud for which the distance to the plane is less than $10cm$, and the orientation of the normal less than $15^{\circ}$. We perform $2000$ iterations and select the plane with the largest number of inliers. The final inliers are defined by a selection criterion: A set of points in the point cloud for which the distance to the plane is less than $20cm$, and the orientation of the normal is less than $30^{\circ}$. If the number of inliers of the plane is higher than $5000$, we add the plane to the set of layout planes and repeat the RANSAC procedure on the remaining set of outliers. If the number is lower, we perform a second stage RANSAC that seeks to find planes corresponding to small layout components. 

In this stage, we set the inlier criterion as follows: A set of points in the point cloud for which the distance to the plane is less than $100cm$, and the orientation of the normal is less than $10^{\circ}$. The same criterion is used for the final selection. If the number of inliers of the plane is higher than $300$, we add the plane to the set of layout planes and repeat the RANSAC procedure on the remaining set of outliers. If the number is lower, we conclude the plane fitting stage. 

\begin{figure*}
    \centering
    \begin{tabular}[t]{ccc}
        \includegraphics[width=0.15\linewidth]{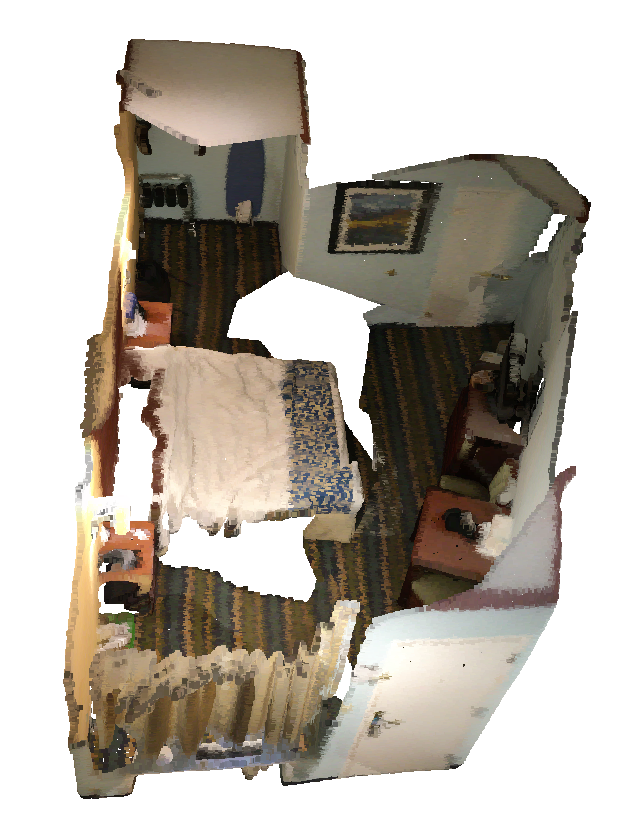} & 
         \includegraphics[width=0.6\linewidth]{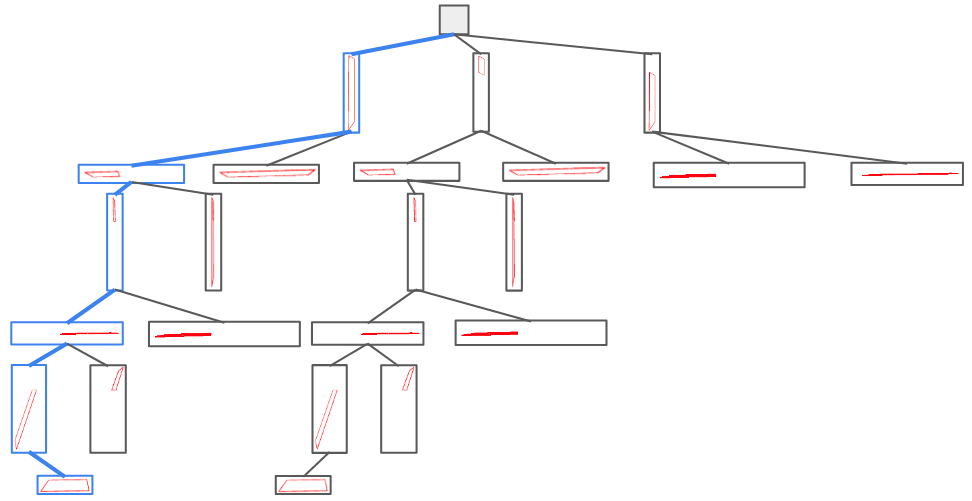} & 
         \includegraphics[trim=0 150 0 0,clip,width=0.18\linewidth]{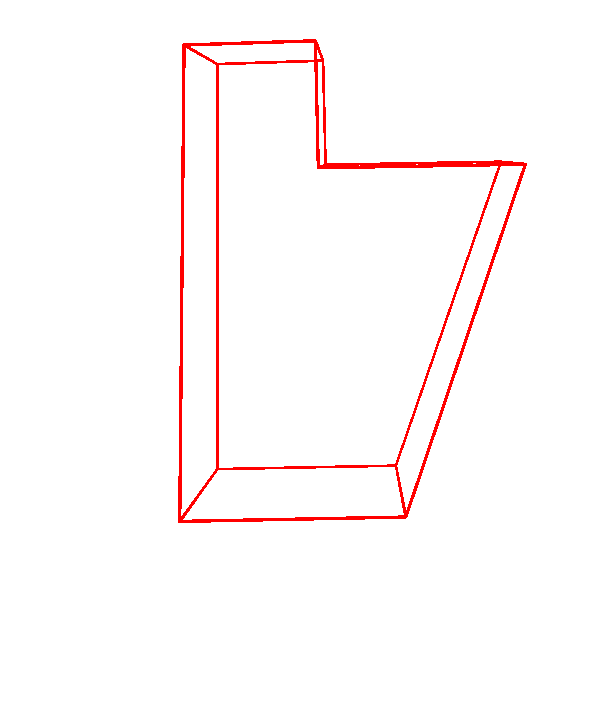} \\
         Input scene & Layout tree after MCSS & Final solution\\
    \end{tabular}
    \caption{The layout proposals are organized into a tree structure such that proposals at the same level of the tree are incompatible to each other but compatible with proposals of their ancestor nodes. Our MCSS approach builds the search tree online and efficiently finds the optimal path, outlined blue, without exploring all candidate solutions.}
    \label{fig:layout_tree}
\end{figure*}

Then, we proceed to define the set of layout proposals by intersecting the layout planes. More exactly, intersections between non-parallel planes triples are candidate corners for the layout. By connecting the vertices that share a pair of layout planes, we get a set of candidate edges. Finally, by connecting the edges that lie on the same layout plane, we extract a set of valid planar polygons for each of the planes. As the SceneCAD dataset contains only scenes with a single floor level, it is enough to perform the search procedure on wall proposals only: the floor polygon can be directly determined afterwards from the walls. This procedure results in a large number of proposals. For  non-cuboid scenes, we obtain between 100 and 1000 proposals, but MCSS can efficiently select the final proposals as shown in Fig.~\ref{fig:layout_tree}. %\shreyasrmk{Should we have a small separate section for the tree visualization?}. \sinisarmk{I think we said enough on that topic in the main paper.}

%%%%%%%%%%%%%%%%%%%%%%%%%%%%%%%%%%%%%%%%%%%%%%%%%%%%%%%%%%%%%%%%%%%%%%%%%%%%%%
%%%%%%%%%%%%%%%%%%%%%%%%%%%%%%%%%%%%%%%%%%%%%%%%%%%%%%%%%%%%%%%%%%%%%%%%%%%%%%

%%%%%%%%%%%%%%%%%%%%%%%%%%%%%%%%%%%%%%%%%%%%%%%%%%%%

\section{Object Proposal Generation}

\begin{figure*}
    \centering
    \includegraphics[trim=20 20 00 20,clip,width=0.8\linewidth]{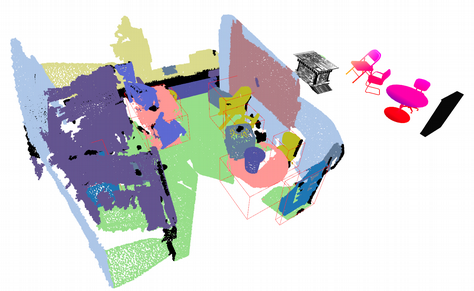}
    \caption{An example synthetic point cloud used for training the network which generates the object proposals. The CAD models corresponding to objects are shown on the right.}
    \label{fig:rendscene}
\end{figure*}

The synthetic point clouds are generated using the ShapeNet~\cite{shapenet} CAD models and the ScanNet~\cite{dai2017scannet} dataset. More specifically, we use the instance annotations of ScanNet and replace the point cloud corresponding to each object with a random CAD model from the same category. The complete scenes with the replaced CAD models are rendered into each of the perspective views using the camera poses and are then reprojected back to 3D. This introduces the incompleteness to the synthetic point cloud due to object occlusions. Furthermore, we also introduce depth holes on the rendered depth maps before reprojecting to 3D to make the point clouds more realistic. Fig.~\ref{fig:rendscene} shows an example of a synthetic scene.

\begin{figure*}
    \centering
    \includegraphics[trim=0 80 00 80,clip,width=0.8\linewidth]{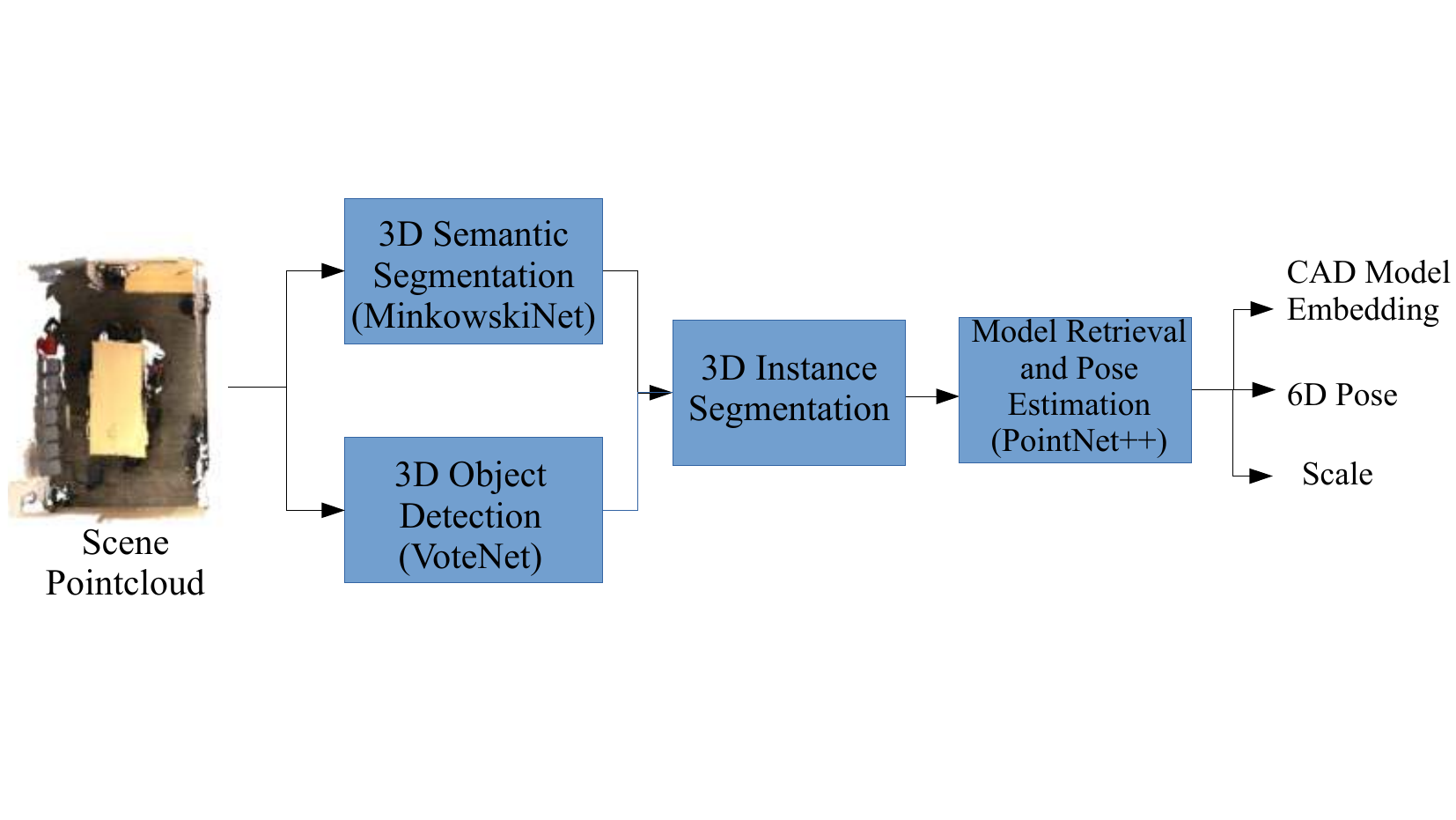}
    \caption{{\bf Object proposals generation pipeline.} We obtain 3D instance segmentation of the input point cloud using the outputs of MinkowskiNet~\cite{choy20194d} and Votenet~\cite{qi2019deep}. We then retrieve multiple CAD models proposals and their corresponding pose+scale for each object instance using a PointNet++ network, which is trained using synthetic data.}
    \label{fig:obj_prop_gen}
\end{figure*}

As explained in Section~4.3 of the main paper and shown in Fig.~\ref{fig:obj_prop_gen}, we use VoteNet~\cite{qi2019deep} and MinkowskiNet~\cite{choy20194d} to extract the point cloud of each object in the scene. A PointNet++ based network trained on the synthetic point clouds is used for object model retrieval and pose estimation. The model retrieval is performed by regressing the embeddings which are obtained by training a PointNet++ auto-encoder on each category of objects. The pose+scale of the object is obtained by regressing the orientation, bounding box center and size. We use the L2 loss with all the embedding and pose+scale parameters.

\begin{figure*}
    \centering
    \includegraphics[trim=0 50 00 50,clip,width=1\linewidth]{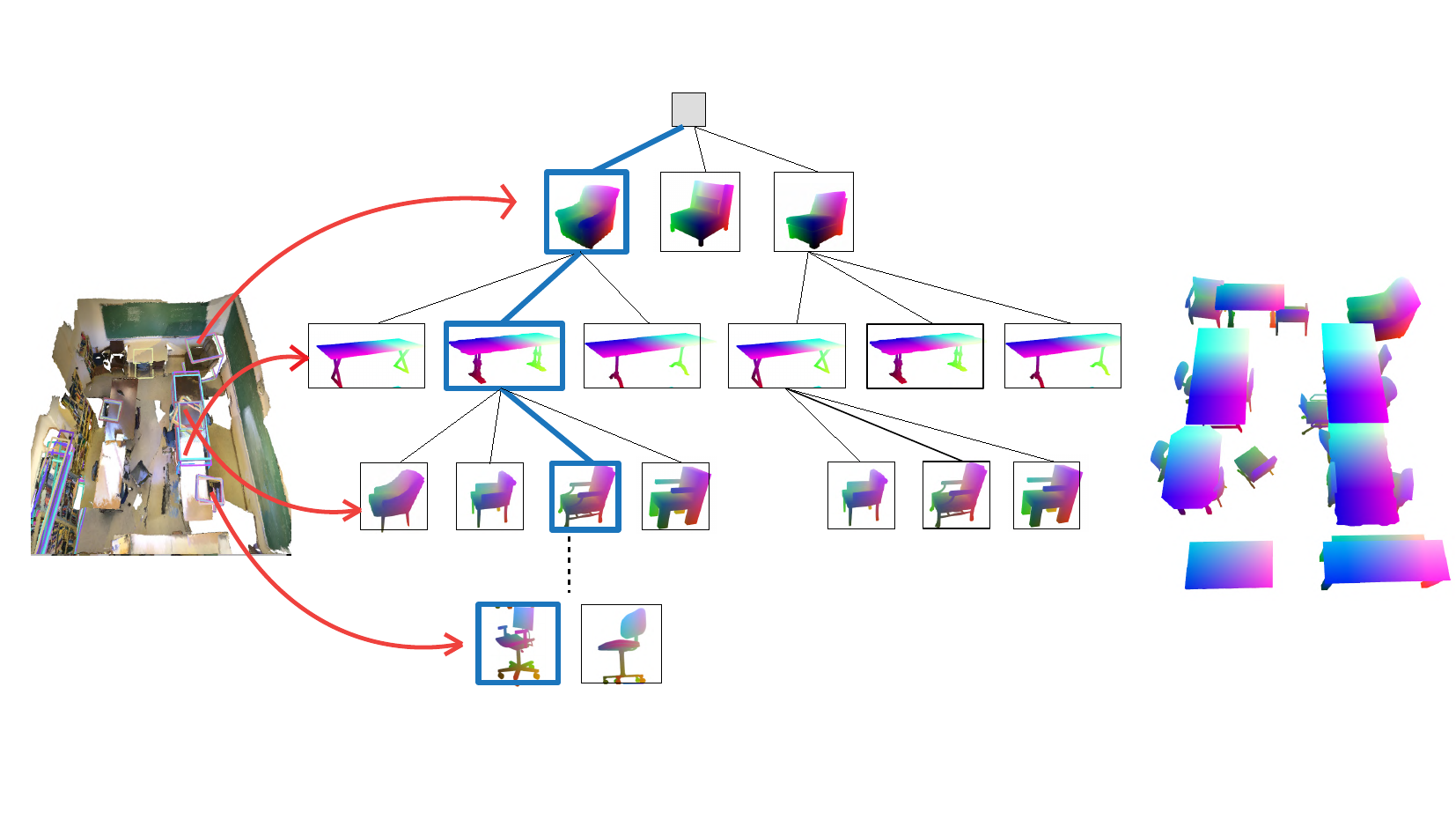}
    RGB-D Input \hspace{4cm} Object Tree after MCSS \hspace{4cm} Final Solution
    \caption{{\bf Visualization of an object tree in MCSS.} At each level of the tree, an object proposal is incompatible with other object proposals at the same level, but compatible with the proposal in the parent node and all its ancestors. MCSS builds the search tree online and finds the optimal path, outlined blue, without exploring all the branches of the tree.}
    %\sinisarmk{We always use "RGB-D". Can you put the sub-captions as latex text, I can see the font is not the same. You can use hspace to put the text where you want it.}
    \label{fig:obj_tree}
\end{figure*}

In Fig.~\ref{fig:obj_tree}, we show the MCSS tree structure for an example scene constructed from several object proposals.

%%%%%%%%%%%%%%%%%%%%%%%%%%%%%%%%%%%%%%%%%%%%%%%%%%%%%%%%%%%%%%%%

\section{MCSS Pseudocode}

MCSS follows the pseudocode for generic MCTS given in Algorithm~\ref{alg:mcts} that is usually used for single-player games. As we explain in the main paper, for the simulation step we can run multiple simulations in practice. For objects, we run 10 simulations in parallel, for layouts we found that running 1 simulation was already enough to achieve robust results.

\input{MCTS_Algo}

% \section{Synthetic Point Cloud Dataset and Object Proposal Generation}

\section{Test Scenes used in Scan2CAD Benchmark}
There are 2 scenes out of 97 scenes we do not consider from the test set while evaluating on the Scan2CAD benchmark, specifically \textit{scene0791\_00} and \textit{scene0793\_00}.
\textit{scene0791\_00} contains multiple floor planes, a special case that we do not address in the object tree, and \textit{scene0793\_00} which contains inconsistent manual annotations as the canonical pose of the \textit{chairs} in the ground truth pool are different.

\section{Computation Times}
For a typical scene with 20 walls and 10 objects, the proposal generation and pre-rendering requires $\sim$15 mins for objects and $\sim$5 mins for layouts. Our MCSS tree search takes 5 mins for 7K iterations on an Intel i7-8700 machine. We would like to point that the proposal generation time especially for objects can be significantly improved by using simplified object models and parallel computations.

% \newpage~\newpage~\newpage~\newpage~\newpage
%%%%%%%%%%%%%%%%%%%%%%%%%%%%%%%%%%%%%%%%%%%%%%%%%%%%%%%%%%%%%%%%%%%%%%%%%%%%%%
%%%%%%%%%%%%%%%%%%%%%%%%%%%%%%%%%%%%%%%%%%%%%%%%%%%%%%%%%%%%%%%%%%%%%%%%%%%%%%

% \clearpage 

\section{Comparisons and Visual Results}
\newcommand{\scannetscene}[1]{\textit{scene#1\_00}}

\subsection{Hill Climbing Baseline}

In addition to the VoteNet baseline for objects~(see Section~5.2 of the main paper), for reference, we also compare our method to a more simple hill climbing optimization algorithm than MCSS for both layouts and objects. At each iteration, the hill climbing algorithm selects the proposal that results in the maximum increase in the scoring function.
It stops when no proposal results in an increase. We consider two different scoring functions for the hill climbing algorithm: 
\begin{itemize}
\item our scoring function $S(\calO)$ used in MCSS~(see Section~4.1 of the main paper). In this case, the selection depends also of the previously selected proposals and the whole images, as the likelihood terms depend on all the image locations. We do not consider proposals that are incompatible with the previously selected proposals.
\item the \textit{fitness} of the proposal~(see Section~4.2.1 of the main paper). In this case, the scoring function depends mainly on the proposal, but we still use the intersection term in cases of objects, and do not consider proposals that are incompatible with the previously selected proposals.
\end{itemize}
The hill climbing algorithm is very simple but provides a local minimum.

More generally, most tree search algorithms will prune parts of the tree based on local heuristics. By contrast, MCTS explores the tree up to the leaves, which allows it to look efficiently for the solution based on a global score.

\subsection{Layout Estimation}

\newcommand{\niceresultwidth}{0.23\linewidth}
\newcommand{\niceresult}[1]{
\includegraphics[width=\niceresultwidth]{figures/qual_supp_layout_jpg/#1_scene.jpg} &
\includegraphics[width=\niceresultwidth]{figures/qual_supp_layout_jpg/#1_baseline.jpg} &
\includegraphics[width=\niceresultwidth]{figures/qual_supp_layout_jpg/#1_our.jpg} &
\includegraphics[width=\niceresultwidth]{figures/qual_supp_layout_jpg/#1_gt.jpg} \\
}

\begin{figure*}
    \centering
    \begin{tabular}{c|c|c|c}
        % \niceresult{scene0030_00}
        RGB-D scan & Annotations from \cite{avetisyan2020scenecad} & MCSS output & Our manual annotations\\
        \niceresult{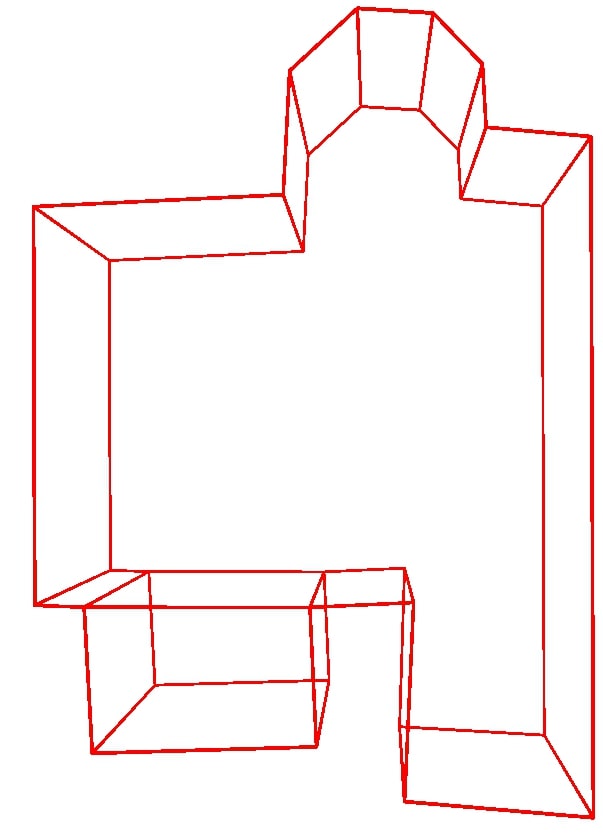} 
        \niceresult{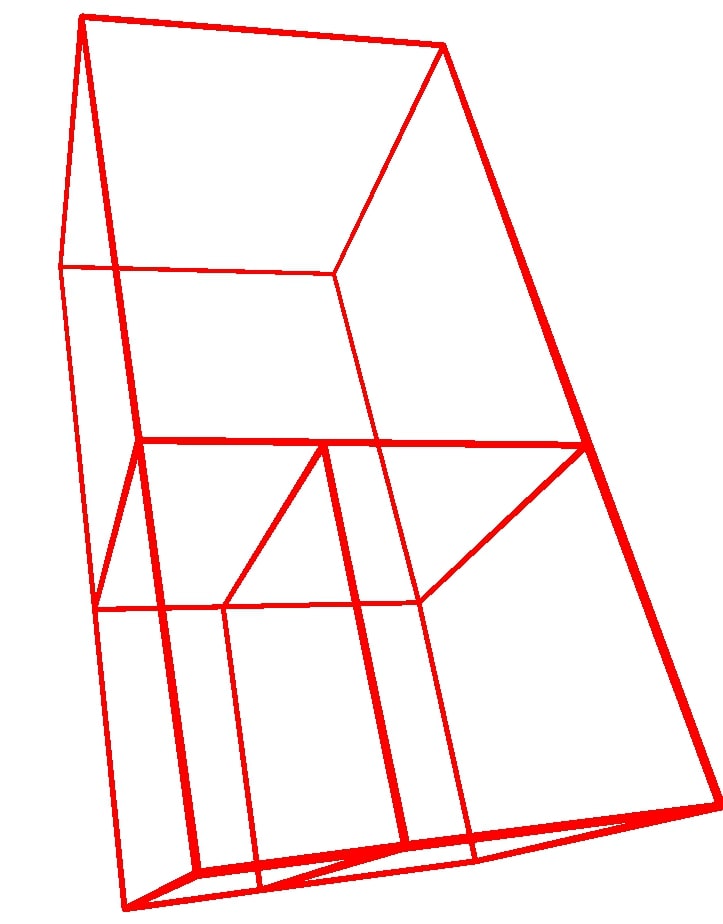} 
        \niceresult{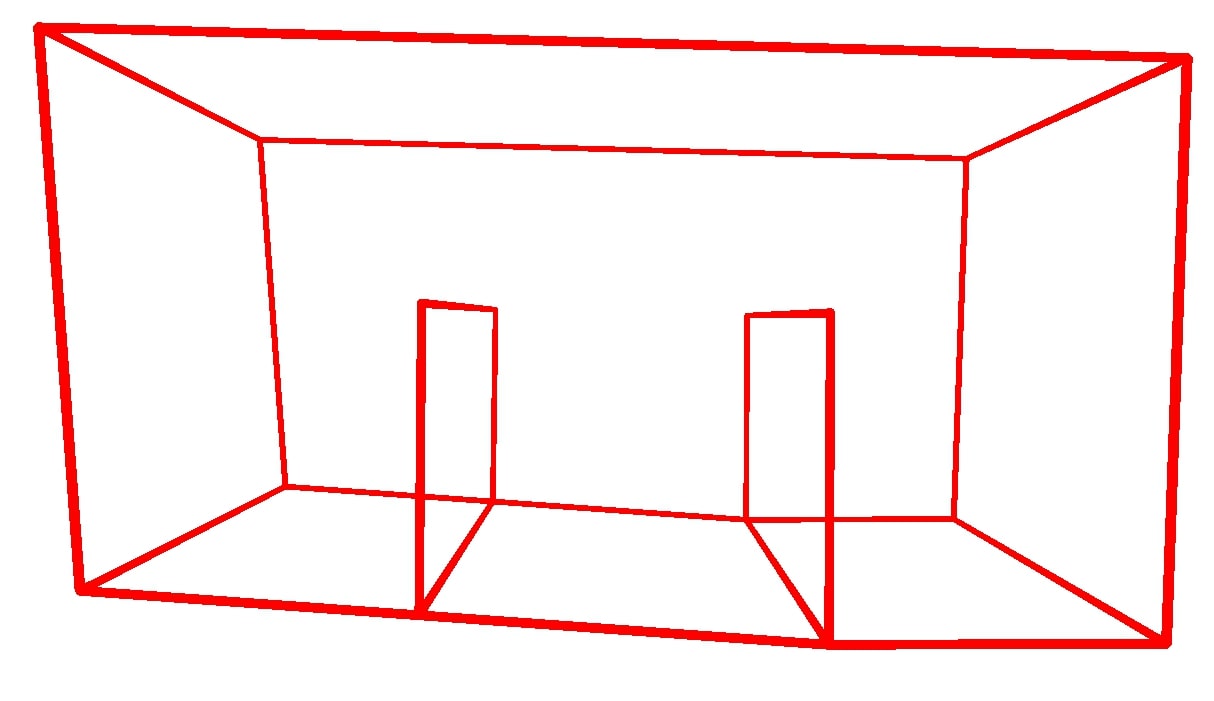} 
        % \niceresult{scene0389_00} %commented to save some space
        \niceresult{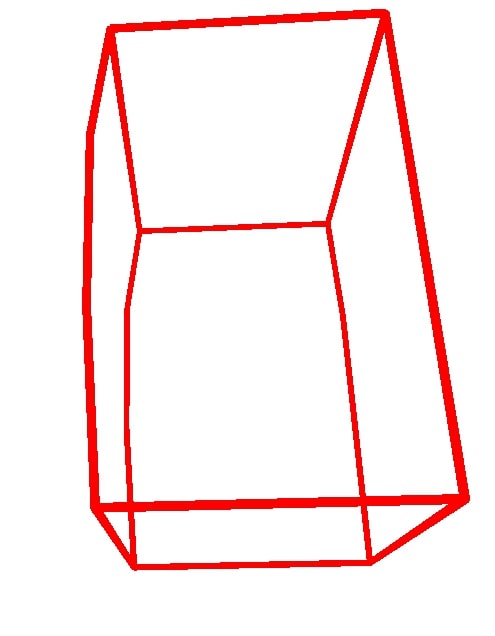} 
        \niceresult{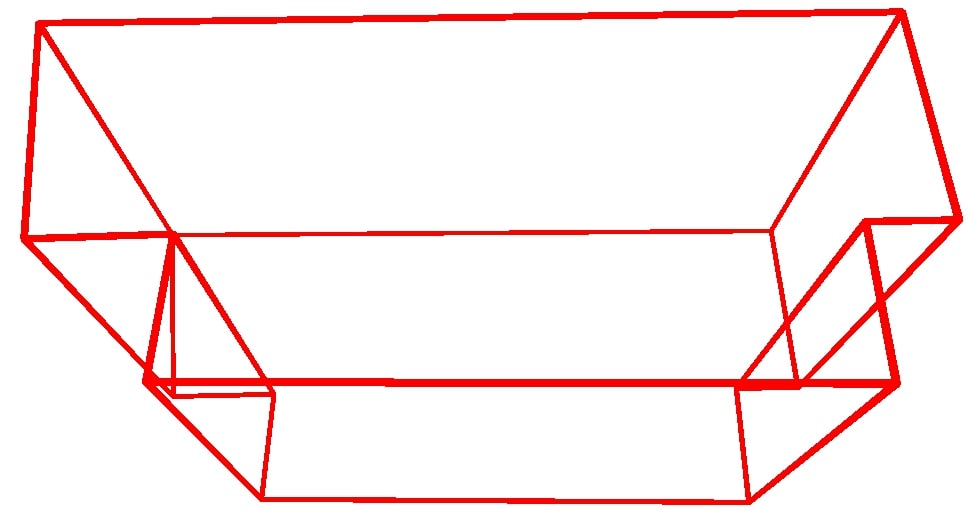}
        % (a) & (b) & (c) & (d) \\
    \end{tabular}
    \caption{RGB-D scans from ScanNet~\cite{dai2017scannet}, existing manual annotations, output of our MCSS approach, and our new manual annotations. Note that we retrieve many details despite the noise and missing data in the scans.}
    \label{fig:layout_qualitative}
\end{figure*}

Fig.~\ref{fig:layout_qualitative} compares the RGB-D scans, the layout annotations from \cite{avetisyan2020scenecad}, the layouts retrieved by our MCSS approach, and our new manual annotations for several representative scenes from the ScanNet dataset~\cite{dai2017scannet}. We show Scenes \scannetscene{0645}, \scannetscene{0046}, \scannetscene{0084}, \scannetscene{0406}, and \scannetscene{0278}. Note that MCSS retrieves detailed layouts, despite noise and missing 3D data.

Fig.~\ref{fig:layout_uphill} shows typical outputs for the hill climbing algorithm. Using our scoring function performs slightly better than simply using the proposals' fitness, however the results are far from perfect as it focuses on the largest components, which may be wrong.

    % \caption{Qualitative results on layout estimation. (a) An RGB-D scan from the ScanNet dataset~\cite{dai2017scannet}. (b) Layout annotations from \cite{avetisyan2020scenecad} lack some details. (c) Layout prediction by our MCSS method. (d) Our manual layout annotations.}

%%%%%%%%%%%%%%%%%%%%%%%%%%%%%%%%%%%%%%%%%%%%%%%%%%%%%%

\subsection{Objects Retrieval and Pose Estimation}

Fig.~\ref{fig:objects_qualitative} compares the RGB-D scans, the 3D pose and model annotations from \cite{Avetisyan_2019_CVPR}, the 3D poses and models  retrieved by our MCSS approach, and the output of the VoteNet baseline~(see Section~5.2 of the main paper) for several representative scenes from the ScanNet dataset~\cite{dai2017scannet}. We show Scenes \scannetscene{0249}, \scannetscene{0549}, \scannetscene{0690}, \scannetscene{0645}, \scannetscene{0342}, and \scannetscene{0518}.

Our method retrieves objects that are not in the manual annotations and sometimes more accurate models: 
See for example the bed in the 5-th row of Fig.~\ref{fig:objects_qualitative}. The VoteNet baseline often fails when the objects are close to each other. 

Fig.~\ref{fig:hill_object} shows the results of hill climbing, compared to the output of MCSS and manual annotations. The hill climbing algorithm tends to choose  large object proposals whenever available, leading to more simplistic solutions that often misses the finer details. Using \textit{fitness} for the scoring function does not consider the occlusions between objects and results in even inferior results.

\section{More Qualitative Results}

\begin{figure*}[t]
    \centering
    \begin{tabular}{cc}
        \includegraphics[trim=80 110 20 80,clip,width=0.5\linewidth]{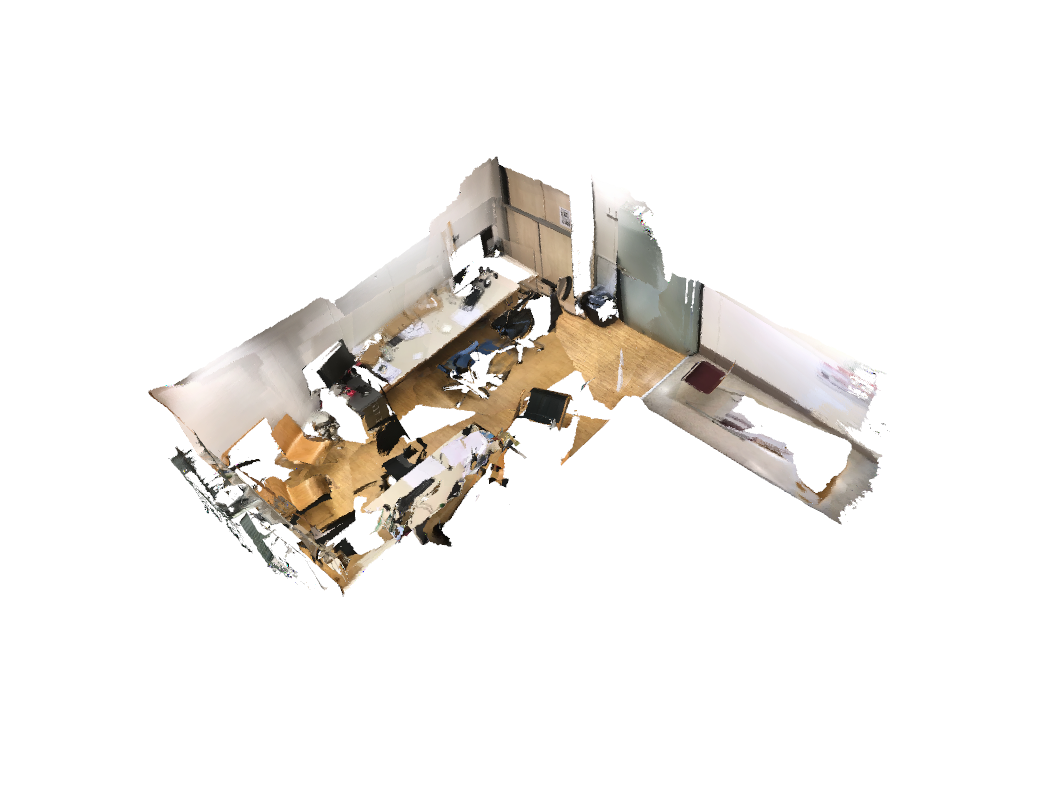} &
\includegraphics[trim=80 110 20 80,clip,width=0.5\linewidth]{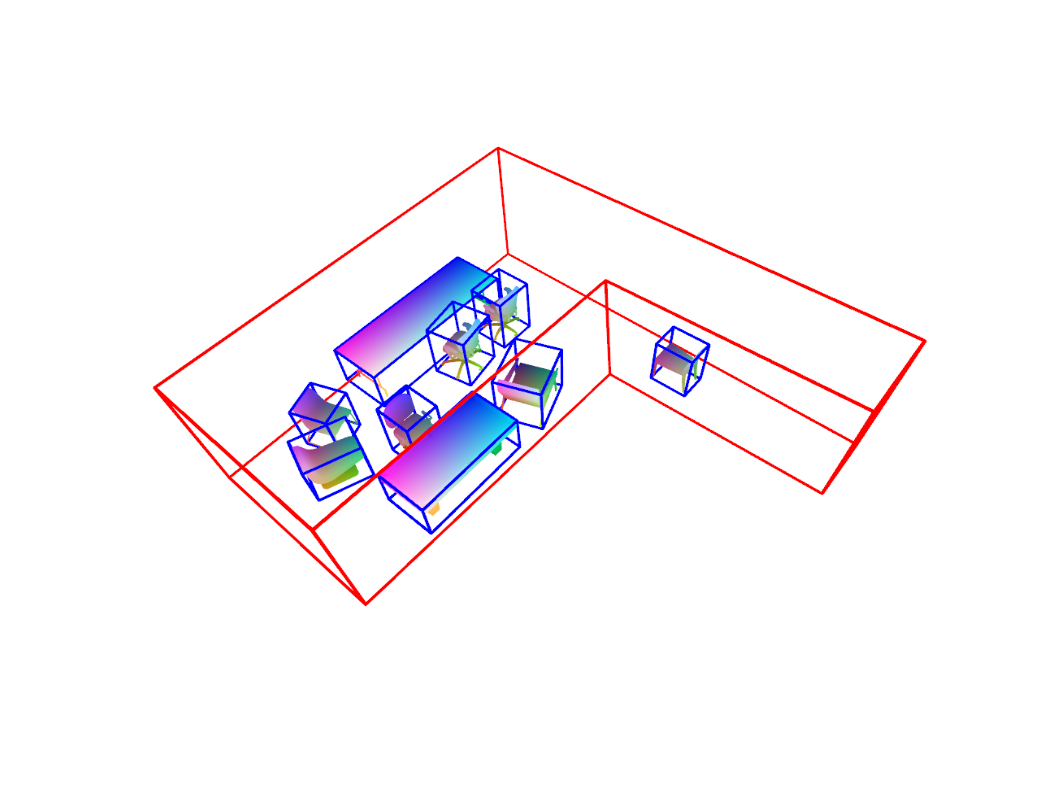} \\

\includegraphics[width=0.45\linewidth]{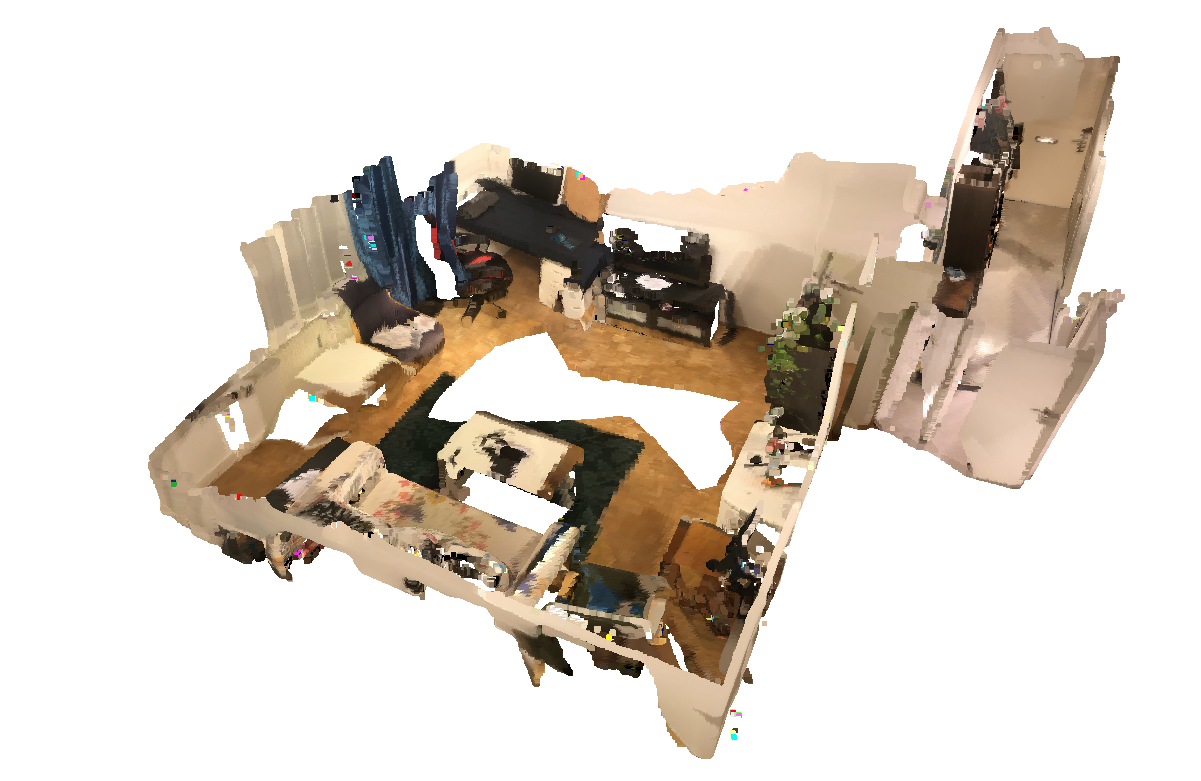} &
\includegraphics[width=0.45\linewidth]{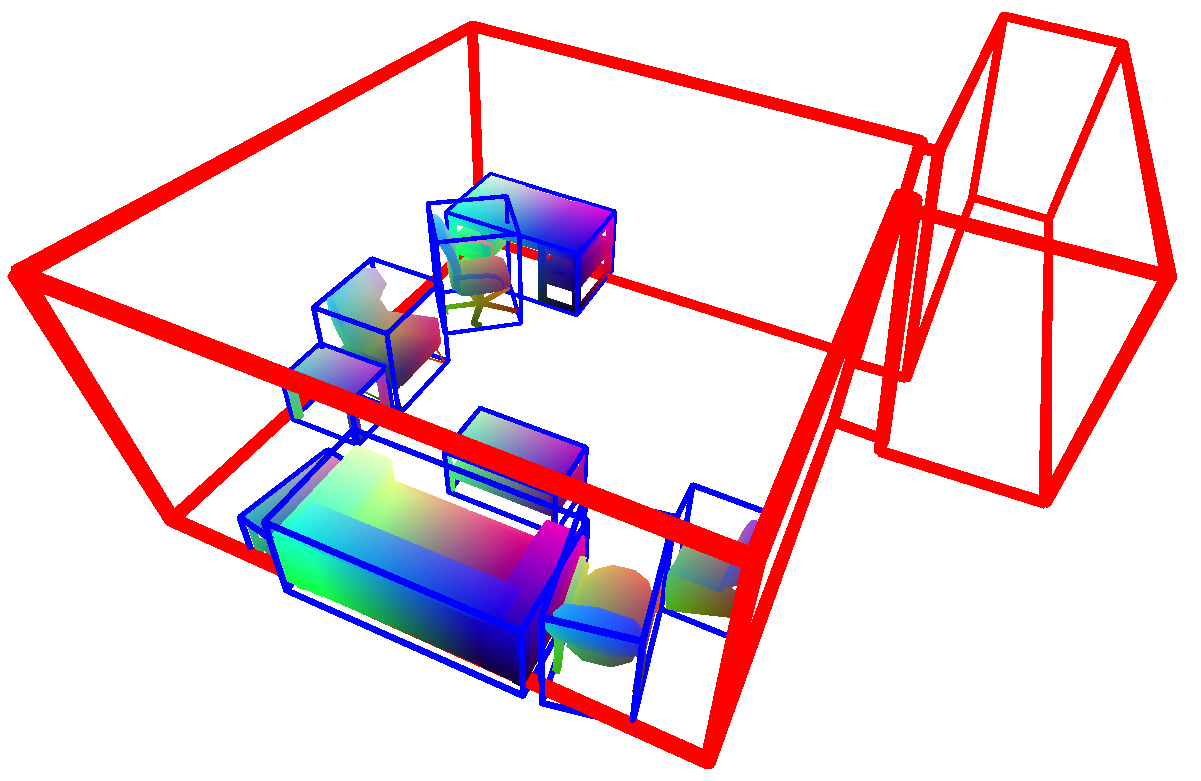} \\

(a) & (b)
    \end{tabular}
    \caption{RGB-D scans of the authors' office and apartment (a) and the automatically retrieved object models from the full ShapeNet dataset and layout (b). Our method generalizes well to RGB-D scans outside the ScanNet dataset.  Note the large areas with missing data, in particular for the layout. }
    \label{fig:office_scan}
\end{figure*}

To show that our method can be applied without retraining nor tuning, we scanned additional scene~(the authors' office and apartment), and applied MCSS. Fig.~\ref{fig:office_scan} shows the scan and the retrieved layouts and objects.

\newcommand{\uphillresultwidth}{0.17\linewidth}
\newcommand{\uphillresult}[1]{
\includegraphics[width=\uphillresultwidth]{figures/uphill_layout_jpg/#1_scene.jpg} &
\includegraphics[width=\uphillresultwidth]{figures/uphill_layout_jpg/#1_uphill_fitness.jpg} &
\includegraphics[width=\uphillresultwidth]{figures/uphill_layout_jpg/#1_uphill_score_fitness.jpg} &
\includegraphics[width=\uphillresultwidth]{figures/uphill_layout_jpg/#1_our.jpg} &
\includegraphics[width=\uphillresultwidth]{figures/uphill_layout_jpg/#1_gt.jpg} \\
}
\begin{figure*}
    \centering
    \begin{tabular}{c|c|c|c|c}
        % \niceresult{scene0030_00}
                   & Hill climbing based & Hill climbing based && \\
        RGB-D scan & on fitness & on our score function & MCSS output & our manual annotations\\
        \uphillresult{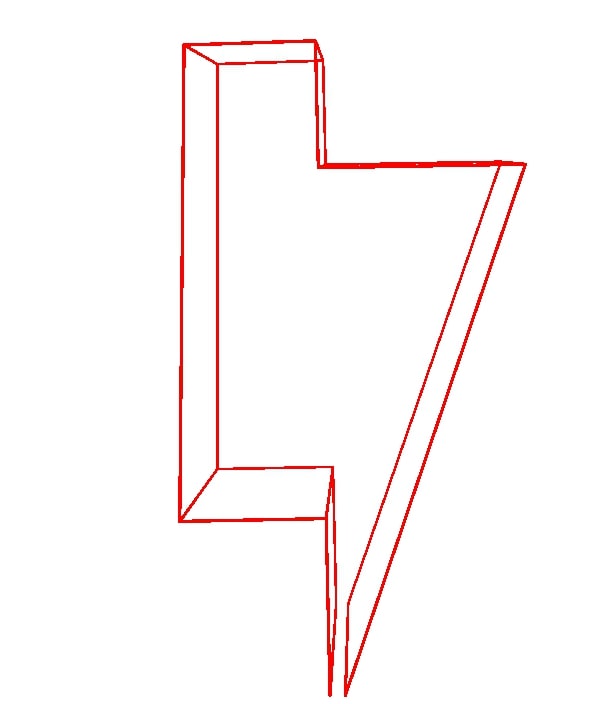} 
        \uphillresult{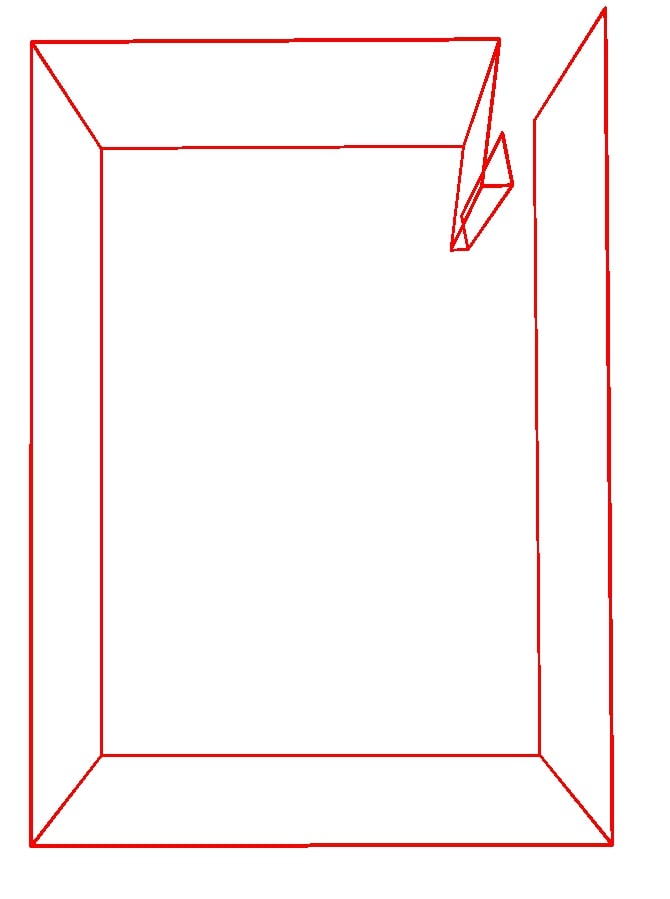} 

        % (a) & (b) & (c) & (d) & (e) \\
    \end{tabular}
    \caption{Typical results of the hill climbing optimization for layout estimation and our results. Using our full scoring function slightly helps but the hill climbing algorithm tends to select large components first and cannot recover when they are incorrect. By contrast, our MCSS approach recovers detailed layouts. }
    \label{fig:layout_uphill}
\end{figure*}

%%%%%%%%%%%%%%%%%%%%%%%%%%%%%%%%%%%%%%%%%%%%%%%%%%%%%%%%%%%%%%%%%%%%%%%%%%%%%%
%%%%%%%%%%%%%%%%%%%%%%%%%%%%%%%%%%%%%%%%%%%%%%%%%%%%%%%%%%%%%%%%%%%%%%%%%%%%%%

\newcommand{\nicerresultwidth}{0.23\linewidth}
\newcommand{\nicerresult}[1]{
\includegraphics[width=\nicerresultwidth]{figures/qual_supp_objects/#1_scene.jpg} &
\includegraphics[width=\nicerresultwidth]{figures/qual_supp_objects/#1_baseline_objects.png} &
\includegraphics[width=\nicerresultwidth]{figures/qual_supp_objects/#1_mcts_objects.png} &
\includegraphics[width=\nicerresultwidth]{figures/qual_supp_objects/#1_scan2cad_objects.png} \\
}
\newcommand{\nicerrresult}[1]{
\includegraphics[width=\nicerresultwidth]{figures/qual_supp_objects_jpg/#1_scene.jpg} &
\includegraphics[width=\nicerresultwidth]{figures/qual_supp_objects_jpg/#1_scan2cad_objects.jpg} &
\includegraphics[width=\nicerresultwidth]{figures/qual_supp_objects_jpg/#1_mcts_objects.jpg} &
\includegraphics[width=\nicerresultwidth]{figures/qual_supp_objects_jpg/#1_baseline_objects.jpg} \\
}
\begin{figure*}
    \centering
    \begin{tabular}{c|c|c|c}
        RGB-D scan & Annotations from \cite{Avetisyan_2019_CVPR} & MCSS output & \textit{VoteNet Baseline}\\

        % \niceresult{scene0030_00}
        % \nicerresult{scene0500_00} 
        \nicerrresult{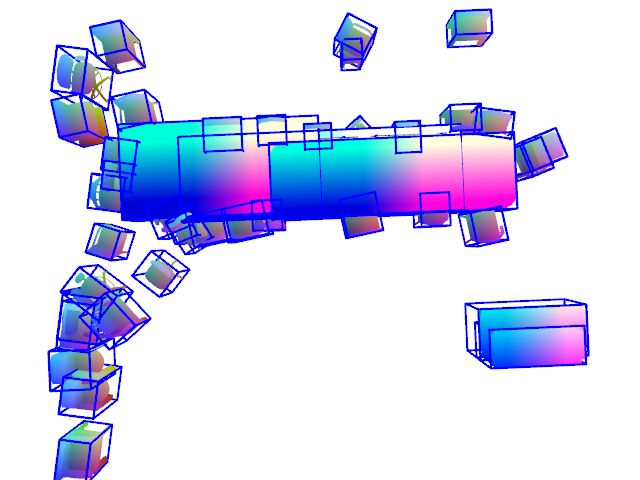} 
        \nicerrresult{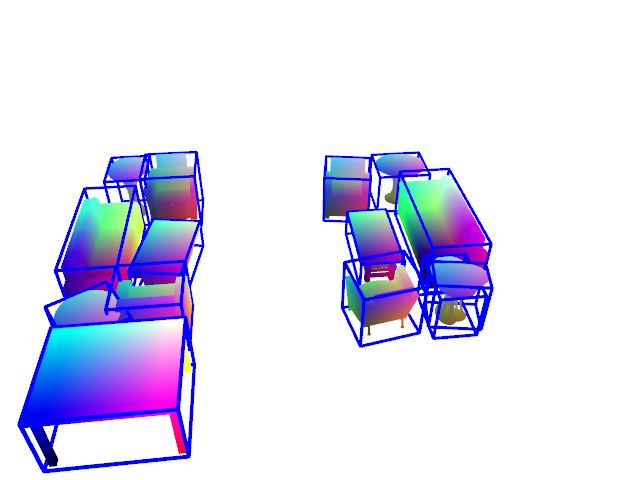} 
        \nicerrresult{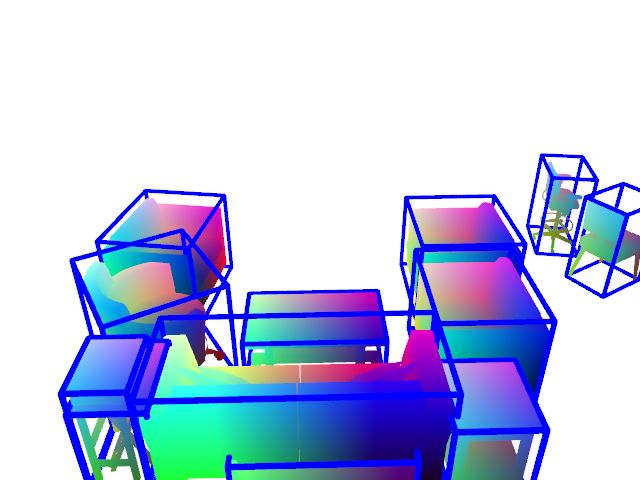} 
        \nicerrresult{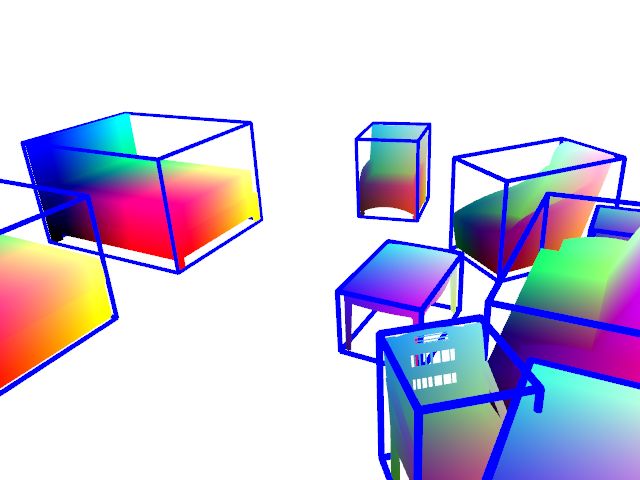} 
        \nicerrresult{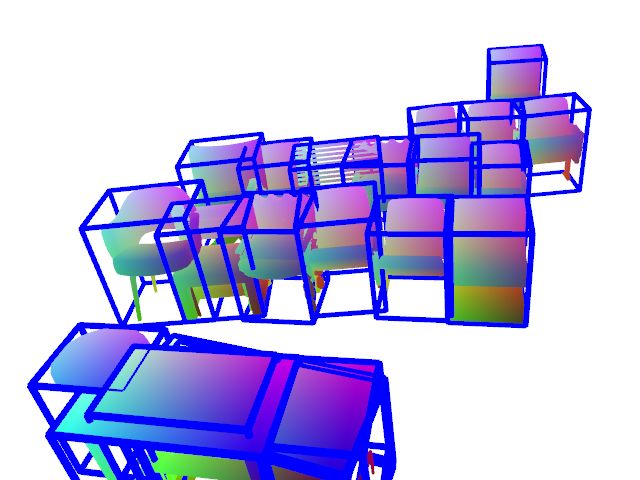} 
        \nicerrresult{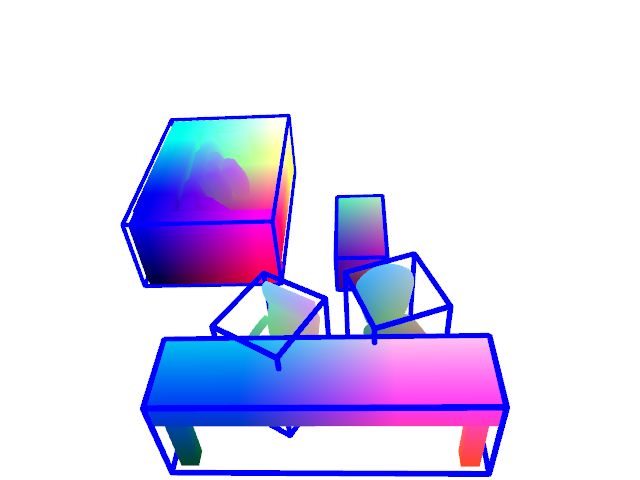} 

        % (a) & (b) & (c) & (d) \\
    \end{tabular}
    \caption{RGB-D scans from ScanNet~\cite{dai2017scannet}, existing manual annotations, output of our MCSS approach, and output of VoteNet for object 3D pose and model retrieval. Note we retrieve objects (shown in red boxes) that are not in the manual annotations, and that VoteNet tends to miss objects or recover an incorrect pose or model when objects are close to each other.}
    % \vspace{cm}
    \label{fig:objects_qualitative}
\end{figure*}

\newcommand{\hcresultwidth}{0.17\linewidth}
\newcommand{\hillclimbresult}[1]{
\includegraphics[width=\hcresultwidth]{figures/qual_supp_objects/#1_scene.png} &\includegraphics[width=\hcresultwidth]{figures/qual_supp_objects/#1_hillClimbing_objects_fitness.png}&
\includegraphics[width=\hcresultwidth]{figures/qual_supp_objects/#1_hillClimbing_objects_global_score.png} &
\includegraphics[width=\hcresultwidth]{figures/qual_supp_objects/#1_mcts_objects.png} &
\includegraphics[width=\hcresultwidth]{figures/qual_supp_objects/#1_scan2cad_objects.png} \\
}
\begin{figure*}
    \centering
    \begin{tabular}{c|c|c|c|c}
                   & Hill climbing based & Hill climbing based &&Scan2CAD  \\
        RGB-D scan & on fitness & on our score function & MCSS output & manual annotations\\
        \hillclimbresult{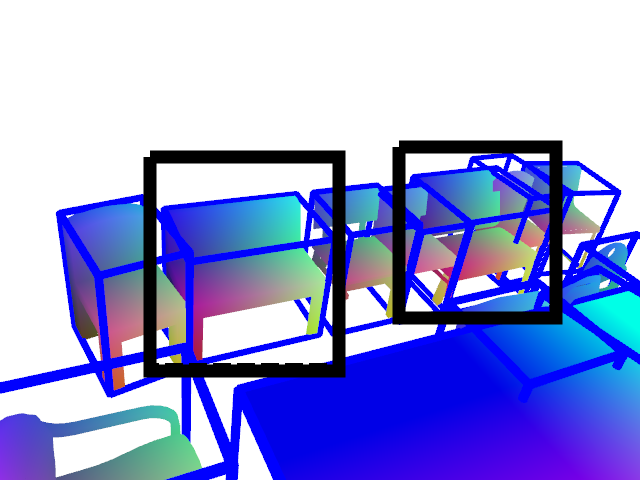} 
        \hillclimbresult{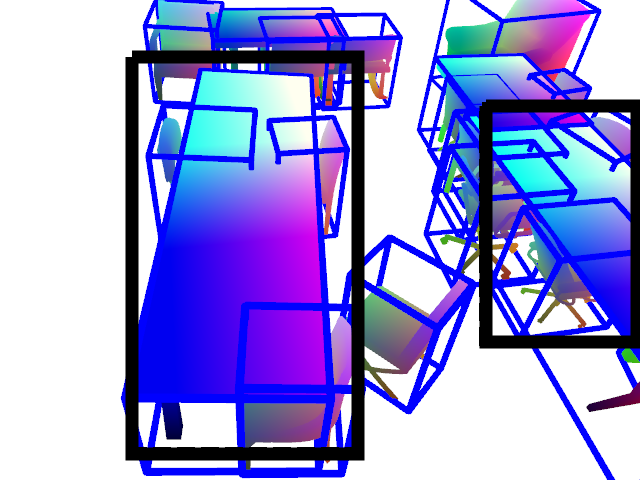} 
        % (a) & (b) & (c) & (d) & (e) \\
    \end{tabular}
    \caption{Typical results of the hill climbing optimization for object pose and model retrieval. The Hill climbing algorithm tends to first focus on large object proposals (shown in black boxes), which may be wrong.}
    \label{fig:hill_object}
\end{figure*}

%%%%%%%%%%%%%%%%%%%%%%%%%%%%%%%%%%%%%%%%%%%%%%%%%%%%%%%%%%%%%%%%%%%%%%%%%%%%%%
%%%%%%%%%%%%%%%%%%%%%%%%%%%%%%%%%%%%%%%%%%%%%%%%%%%%%%%%%%%%%%%%%%%%%%%%%%%%%%

%%%%%%%%%%%%%%%%%%%%%%%%%%%%%%%%%%%%%%%%%%%%%%%%%%%%%%%%%%%%%%%%%%%%%%%%%%%%%%
%%%%%%%%%%%%%%%%%%%%%%%%%%%%%%%%%%%%%%%%%%%%%%%%%%%%%%%%%%%%%%%%%%%%%%%%%%%%%%

%%%%%%%%%%%%%%%%%%%%%%%%%%%%%%%%%%%%%%%%%%%%%%%%%%%%%%%%%%%%%%%

%% file: MCTS_Algo.tex
% \sinisa{or $\calN$ is $\calN_\END$}

\begin{algorithm}[t]
    iters $\leftarrow$ Number of desired runs, best\_moves $\leftarrow \varnothing$\;
    \While{\normalfont{iters} $> 0$}{
        $\calN_\curr \leftarrow \calN_\Root$\;
        reached\_terminal $\leftarrow$ False\;
        \While{not \normalfont{reached\_terminal}}{
            $\calN_\curr \leftarrow \fn{Select}(\calN_\curr)$\;
            \If{$\calN_\curr \text{ is visited for the first time}$}{
              \fn{Expand}($\calN_\curr$)\;
              best\_sim $\!\leftarrow\!\arg\!\displaystyle\max_{\text{sim}} sc(\fn{Simulate}(\calN_\curr,\text{sim}))\!\!\!$\;
              %% best\_sim $\leftarrow$\;
              %%     \Indp$\>\arg\displaystyle\max_{\text{sim}} sc(\fn{Simulate}(\calN_\curr,\text{sim}))$\;\Indm
               $\fn{Update}($best\_sim$)$\;
               \If{$sc($\normalfont{best\_sim}$) > sc($best\_moves$)$}{
                  best\_moves $\leftarrow$ moves of best\_sim \;
               }
               reached\_terminal $\leftarrow$ True\;
            }
            \ElseIf{$\calN_\curr$ is terminal}{
            reached\_terminal $\leftarrow$ True \;
            }
            % \sinisa{
            % \ElseIf{$\calN_\curr$ is terminal}{
            % \sinisarmk{Only if score has some randomness}
            % $selected\_moves$ $\leftarrow$  $\fn{GetMoves}(\calN_\curr)$\;
            %   $\fn{Update}($best\_sim$)$\;
            %   \If{$sc($selected\_moves$) > sc($best\_moves$)$}{
            %       $best\_moves$ $\leftarrow$ $selected\_moves$ \;
            %   }
            % reached\_terminal $\leftarrow$ True\;
            % }}
        }
        iters $\leftarrow$ iters - 1\;
    }
    \return best\_moves\;

  \caption{Generic MCTS for non-random single-player games}\label{alg:mcts}
\end{algorithm}

%% \begin{algorithm}
%%     iters $\leftarrow$ Number of desired runs, best\_moves $\leftarrow \varnothing$\;
%%     \While{iters $> 0$}{
%%         $\calN_\curr \leftarrow \calN_\Root$\;
%%         reached\_a\_new\_node $\leftarrow$ False\;
%%         \While{not reached\_a\_new\_node}{
%%             $\calN_\curr \leftarrow \fn{Descend}(\calN_\curr)$\;
%%             \If{$\calN_\curr \text{ is visited for the first time}$}{
%%               best\_sim $\leftarrow$\;
%%                   \Indp$\>\arg\displaystyle\max_{\text{sim}} sc(\fn{OneSimulation}(\calN_\curr,\text{sim}))$\;\Indm
%%                $\fn{UpdateVisitedNodes}($best\_sim$)$\;
%%                \If{$sc($best\_sim$) > sc($best\_moves$)$}{
%%                   best\_moves $\leftarrow$ \sinisa{moves of best\_sim} \;
%%                }
%%                reached\_a\_new\_node $\leftarrow$ True
%%             }
%%         }
%%         iters $\leftarrow$ iters - 1\;
%%     }
%%     \return best\_moves\;
%%     \BlankLine
%%     \Def{\fn{Descend}$(\calN)$}{
%%       blah\;
%%     }
%%     \Def{\fn{OneSimulation}\normalfont{($\calN$, sim)}}{
%%       blah\;
%%     }
%%     \Def{\fn{UpdateVisitedNodes}\normalfont{(sim)}}{
%%       blah\;
%%     }
%%     \caption{generic MCTS}\label{alg:mcts}
%% \end{algorithm}